%% file: main.tex
\documentclass{article}
\usepackage{iclr2027_conference,times}
\usepackage{scad_arxiv}

\input{math_commands.tex}

\usepackage{hyperref}
\usepackage{xurl}
\usepackage{wrapfig}
\usepackage{booktabs,tabularx}
\usepackage[table]{xcolor}
\usepackage{graphicx}
\usepackage{fontawesome5}
\usepackage{twemojis}
\usepackage{tcolorbox}
\usepackage{algpseudocode}
\usepackage[protrusion=false,expansion=true]{microtype}
\newenvironment{scadblock}{\par\begin{samepage}}{\par\end{samepage}\penalty0}
\DeclareFontFamily{U}{wasy}{}
\DeclareFontShape{U}{wasy}{m}{n}{<-> wasy10}{}
\DeclareUnicodeCharacter{266B}{{\usefont{U}{wasy}{m}{n}\symbol{15}}}

\newcount\scadreservedpage
\newcommand{\scadreservepage}{\ifnum\value{page}>\scadreservedpage
        \global\scadreservedpage=\value{page}\enlargethispage{-9.08pt}\fi}
\newcommand{\scadpagebreak}{\par\pagebreak\scadreservepage}
\AddToHook{cmd/bibitem/before}{\scadreservepage}

\definecolor{scadtitle}{RGB}{13,133,136}
\newcommand{\scadtitleaccent}[1]{\textcolor{scadtitle}{\textbf{\upshape #1}}}
\newcommand{\scadtitleicon}{\raisebox{-0.12em}{\includegraphics[height=0.99em]{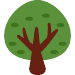}}\hspace{0.2em}}
\title{\raggedright\hyphenpenalty=10000\exhyphenpenalty=10000
\texorpdfstring{\scadtitleicon\scadtitleaccent{SCAD}:
\scadtitleaccent{S}tructured \scadtitleaccent{C}redit
\scadtitleaccent{A}ssignment and
\scadtitleaccent{D}istillation for Long-Horizon Agents}
{SCAD: Structured Credit Assignment and Distillation for Long-Horizon Agents}}
\newcommand{\scadhomepageurl}{https://shangyangwu2002.github.io/scad/}
\newcommand{\scadhomepagelink}{\ifx\scadhomepageurl\empty
        \textcolor{black!50}{\faGlobe\ Homepage}\else
        \href{\scadhomepageurl}{\faGlobe\ Homepage}\fi}
\newcommand{\scadresourcelinks}{\begingroup
    \hypersetup{pdfborder={0 0 0}}\color{scadtitle}\scadhomepagelink\qquad
    \href{https://github.com/shangyangwu2002/SCAD}{\faGithub\ Code}\qquad
    \href{https://huggingface.co/shangyangwu2002/SCAD}{\faCube\ Model}\qquad
    \href{https://huggingface.co/datasets/shangyangwu2002/SCAD-Corpus}{\faDatabase\ Data}\endgroup}
\author{{\normalsize
    \textbf{\mbox{Shangyang Wu\textsuperscript{1}}\quad
    \mbox{Shuai Zhao\textsuperscript{1,$\dagger$}}\quad
    \mbox{Ziyue Zhu\textsuperscript{1}}\quad
    \mbox{Jinyang Wu\textsuperscript{2}}\quad
    \mbox{Anh Tuan Luu\textsuperscript{3}}\quad
    \mbox{Haoran Luo\textsuperscript{3,$\dagger$}}}\\[7pt]
    \mbox{\textsuperscript{1}Beijing University of Posts and Telecommunications}\quad
    \mbox{\textsuperscript{2}Tsinghua University}\quad
    \mbox{\textsuperscript{3}Nanyang Technological University, Singapore}}\\[4pt]
    {\normalsize\ttfamily
    \{\href{mailto:wushangyang@bupt.edu.cn}{wushangyang},\href{mailto:zhaoshuaiby@bupt.edu.cn}{zhaoshuaiby}\}@bupt.edu.cn\quad
    \href{mailto:haoran.luo@ntu.edu.sg}{haoran.luo@ntu.edu.sg}}\\[7pt]
    \makebox[\linewidth][c]{\scadresourcelinks}}
\hypersetup{
    pdftitle={SCAD: Structured Credit Assignment and Distillation for Long-Horizon Agents},
    pdfauthor={Shangyang Wu, Shuai Zhao, Ziyue Zhu, Jinyang Wu, Anh Tuan Luu, Haoran Luo},
    pdfsubject={Structured credit assignment and subtask-local on-policy distillation for long-horizon agents},
    pdfkeywords={SCAD, long-horizon agents, credit assignment, on-policy distillation},
    colorlinks=false,
    pdfborder={0 0 1},
    linkbordercolor={1 0 0},
    citebordercolor={0 1 0},
    urlbordercolor={0 1 1}
}

\begin{document}
\flushbottom
\setlength{\parskip}{5pt plus 2pt minus 0.5pt}
\scadreservepage
\microtypesetup{expansion=false}
\begingroup
\microtypesetup{expansion=true}
\fussy
\setlength{\emergencystretch}{0.5em}
\maketitle
\begingroup
\renewcommand{\thefootnote}{\fnsymbol{footnote}}
\footnotetext[2]{Corresponding authors.}
\endgroup

\begin{abstract}
Training long-horizon agents to solve complex tasks requires effective supervision over extended interaction sequences.
However, sparse terminal rewards obscure intermediate contributions, while on-policy distillation can lose informative teacher guidance as student-generated histories grow.
To address this problem, we introduce \textbf{SCAD}, which organizes interactions into planning and bounded subtask execution, distills execution in local contexts, and refines planning credit through cross-rollout subtask prefix trees, with planning receiving full terminal credit and execution receiving positive terminal credit and teacher guidance.
Across all evaluated benchmarks, SCAD improves macro-average accuracy over the strongest training baseline by 4.48 percentage points for text tasks and 4.19 points for multimodal tasks.
SCAD effectively combines outcome-based credit assignment with teacher-guided distillation to improve planning and execution in long-horizon agents.
\end{abstract}
\vspace{\dimexpr\baselineskip-17pt\relax}

\section{Introduction}
\label{sec:introduction}

\begin{wrapfigure}{r}{0.50\textwidth}
    \centering
    \vspace{-\intextsep}
    \includegraphics[width=\linewidth,height=0.869144981\linewidth]{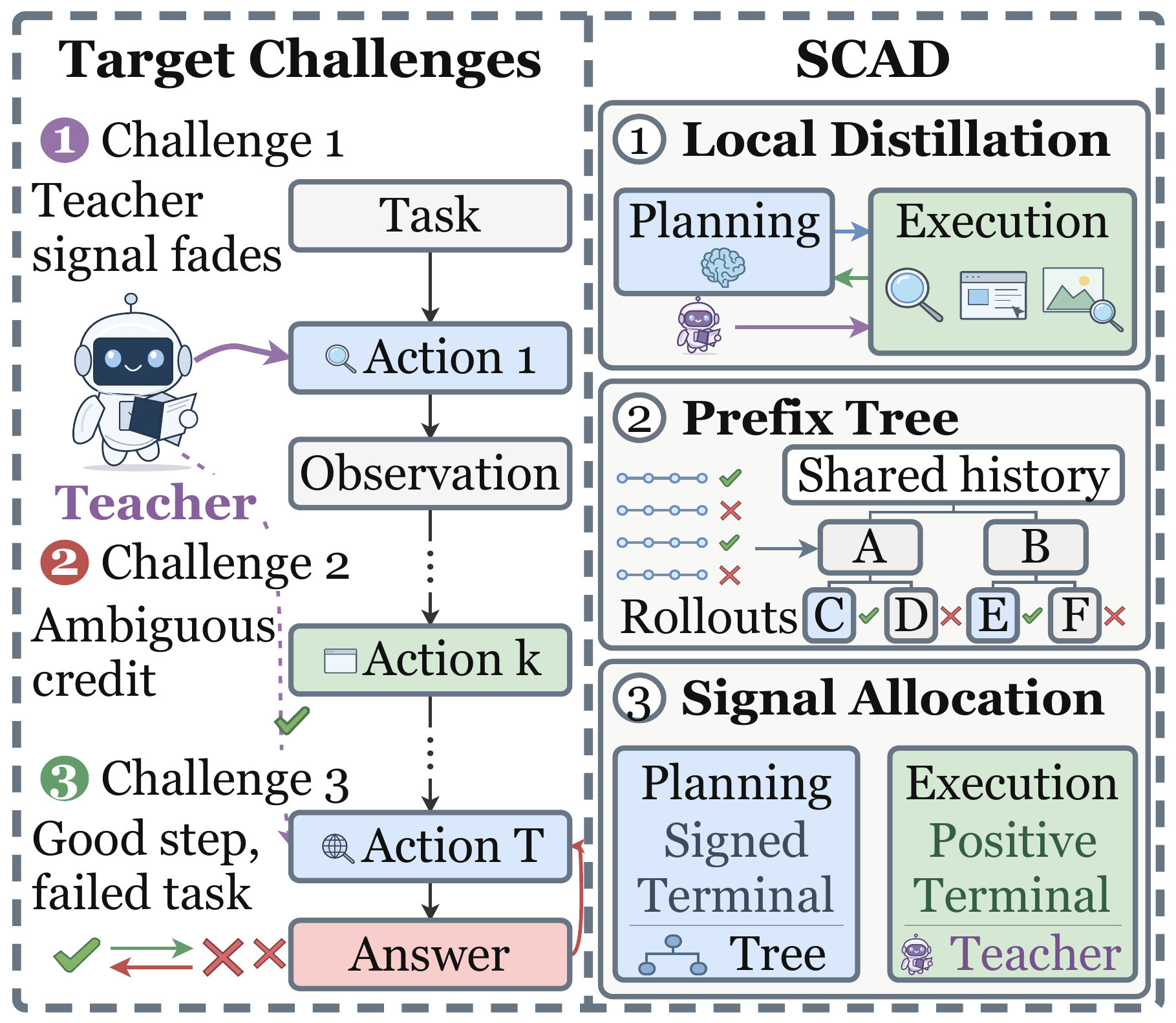}
    \caption{Target challenges and SCAD}
    \label{fig:scad-overview}
\end{wrapfigure}

Language model agents use tools to interact with external environments and accomplish tasks \citep{nakano2022webgptbrowserassistedquestionansweringhuman,yao2022react}. Long-horizon agents solve complex real-world tasks through many interdependent decisions \citep{sun2026scaling}. Earlier actions shape later information and options, requiring agents to coordinate decisions toward an overall goal \citep{wu2025webdancer}. Multimodal tasks further require connecting textual information with visual evidence \citep{li2026mmbrowsecomp}. Success depends on both choosing productive directions and executing individual actions effectively \citep{peng2026hiper}. Training agents to acquire these capabilities requires supervision that remains useful throughout an extended sequence of decisions.

Long-horizon agent training uses reinforcement learning (RL) and on-policy distillation (OPD); the latter builds on supervision at learner-visited states \citep{pmlr-v15-ross11a,pmlr-v89-czarnecki19a}. RL emphasizes task outcomes, while OPD supplies teacher guidance for intermediate behavior. Context Folding and CoDA manage histories through subtasks \citep{sun2026scaling,10.1145/3773966.3777986}. HiPER separates planning and execution advantages; GiGPO groups shared states for finer credit \citep{peng2026hiper,feng2025groupingroup}. Tree-GRPO uses tree-search rollouts to induce implicit step credit \citep{ji2026tree}. MiniLLM uses reverse KL for language-model distillation \citep{gu2024minillm}. TCOD increases trajectory depth to stabilize multi-turn OPD \citep{wang2026exploring}. ATOD combines RL and distillation through annealed weights and turn-level supervision \citep{tan2026atodannealedturnawareonpolicy}. These methods organize supervision at different scales.

\scadpagebreak
\begin{figure}[t]
    \centering
    \raisebox{-\fboxrule}{\includegraphics[width=\linewidth]{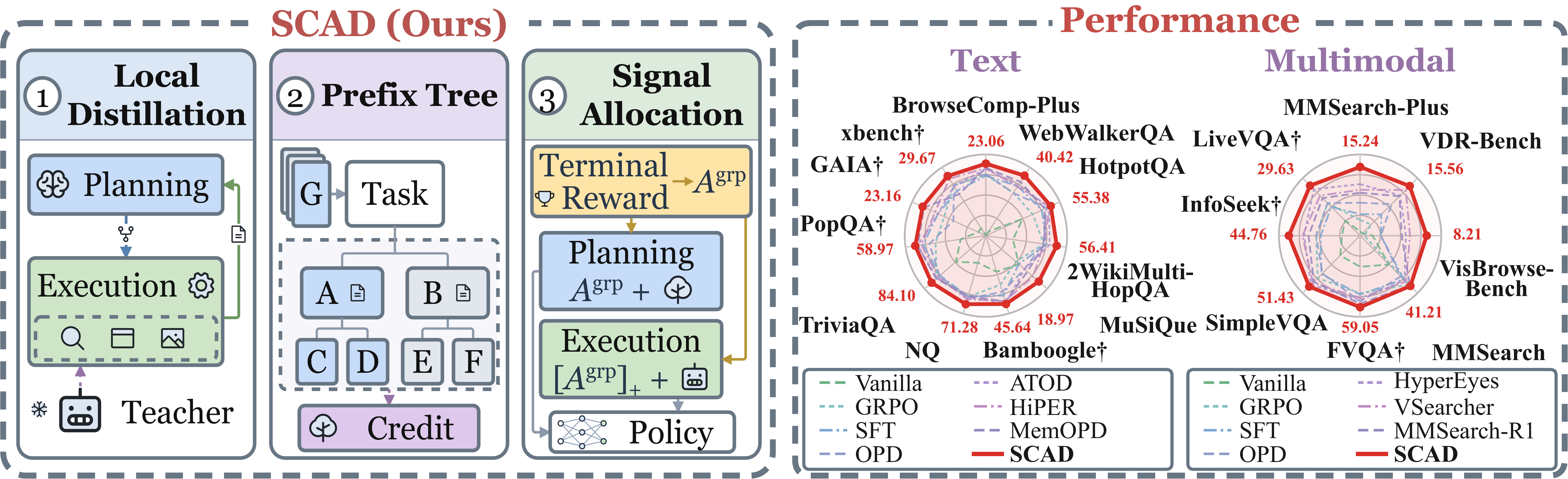}}
    \caption[SCAD overview and performance on long-horizon agent benchmarks.]{SCAD training overview (left) and benchmark accuracy (\%) across text and multimodal tasks (right; SCAD in red). Axes use separate zero-based scales; $\dagger$ marks out-of-distribution datasets.}
    \label{fig:scad-performance}
\end{figure}

However, three challenges remain. \textbf{(i) Maintaining useful teacher supervision over long interaction histories.} Teacher feedback can lose discriminative value along student-generated histories \citep{li2026rethinking,liu2026prefixteachsuffixfade}. \textbf{(ii) Ambiguous credit under long-range dependencies.} Terminal outcomes depend on preceding history and subsequent actions, obscuring individual decision contributions and complicating cross-rollout comparisons \citep{feng2025groupingroup,he2026hierarchyofgroups}. \textbf{(iii)~Coordinating task-level rewards and local learning.} Outcome rewards guide task success, while teacher feedback guides local learning. Uniform mixing can let negative terminal feedback suppress useful teacher guidance for correctly executed actions in failed trajectories \citep{hubotter2026reinforcement}.

To address these challenges, we propose \textbf{SCAD}, a framework for training long-horizon agents through three complementary designs. \textbf{(i) Subtask-local distillation.} We distill bounded subtask execution in local contexts to limit cross-subtask history accumulation, helping sustain useful teacher feedback at later stages. Teacher feedback is computed under the same local context used by the student. \textbf{(ii) Cross-rollout planning credit.} We construct subtask--report prefix trees from completed rollouts and compare terminal returns under matched histories to estimate relative credit for individual planning decisions. \textbf{(iii)~Structured signal allocation.} Planning retains signed terminal and tree credit, while execution receives positive terminal credit and local teacher feedback. This allocation preserves teacher guidance in failed trajectories and reinforces execution associated with task success.

We evaluate SCAD on text-based and multimodal long-horizon benchmarks against RL, OPD, and hybrid baselines. SCAD improves overall macro-average accuracy over the strongest training baseline by 4.48 and 4.19 percentage points, respectively. Efficiency analysis shows lower training time than flat GRPO + OPD. Ablations and diagnostics examine component contributions, supervision quality across interaction depths, planning credit, and the link between local execution and global success.

\section{Related Work}
\label{sec:related-work}

\textbf{Long-Horizon Agents and Architectures.} Long-horizon agents gather textual and visual evidence \citep{nakano2022webgptbrowserassistedquestionansweringhuman,jin2025searchr,wu2025webdancer,wu2025mmsearchr1incentivizinglmmssearch,zhang2026vsearcherlonghorizonmultimodalsearch}. AgentFlow separates planning and execution \citep{li2026intheflow}; Context Folding and CoDA bound execution contexts \citep{sun2026scaling,10.1145/3773966.3777986}. AgentFold and IterResearch compress histories through folding or workspace reconstruction \citep{ye2026agentfold,chen2026iterresearch}. SCAD extends context isolation to teacher supervision, distilling execution locally while planning retains task history for coordination.

\textbf{Credit Assignment and Policy Distillation.} Credit assignment uses return redistribution \citep{NEURIPS2019_16105fb9}, intermediate rollouts \citep{3780338.3781496,hou-etal-2025-treerl,ji2026tree}, shared-state or prefix comparisons \citep{feng2025groupingroup,he2026hierarchyofgroups,wang2026rtmcsteplevelcreditassignment,tran-etal-2026-exploiting}, and hierarchical or role-based advantages \citep{peng2026hiper,xu2026triageroletypedcreditassignment}. Distillation \citep{hinton2015distillingknowledgeneuralnetwork,rusu2016policydistillation} uses on-policy learning \citep{agarwal2024onpolicy,gu2024minillm,ko2025distillm}, temporal curricula \citep{wang2026exploring}, and memory alignment \citep{liu2026memopdonpolicydistillationmemory}. Kickstarting \citep{schmitt2018kickstartingdeepreinforcementlearning} and hybrid methods \citep{hubotter2026reinforcement,tan2026atodannealedturnawareonpolicy,lin2026onpolicydistillationverifiablereward} combine RL and distillation. SCAD derives planning credit from subtask--report prefixes, blocks negative terminal credit to execution, and retains local teacher guidance.

\scadpagebreak
\begingroup
\setlength{\textfloatsep}{8pt plus 1pt minus 1pt}
\setlength{\parskip}{3pt plus 1pt minus 0.5pt}
\section{Methodology}
\label{sec:methodology}
\input{tables/method_comparison}

\input{sections/methodology}

\scadpagebreak
\endgroup
\endgroup
\input{sections/experiments}

\scadpagebreak
\microtypesetup{expansion=true}
\bibliographystyle{scad_references}
\bibliography{references}

\clearpage
\appendix
\setlength{\parskip}{3pt plus 3pt minus 0.5pt}
\AddToHook{para/before}[scad-appendix-reserve]{\scadreservepage}
\interlinepenalty=10000
\fussy
\setlength{\emergencystretch}{0.5em}
\input{sections/additional_experimental_results}
\clearpage\scadreservepage
\input{sections/prompts}
\clearpage\scadreservepage
\input{sections/proposition_proofs}
\scadpagebreak
\input{sections/algorithm_details}
\clearpage\scadreservepage
\input{sections/benchmark_details}
\clearpage\scadreservepage
\input{sections/resource_sources}
\clearpage\scadreservepage
\begingroup\raggedbottom
\input{sections/case_study}
\input{sections/limitations_future_work}

\par\scadreservepage\newpage
\endgroup
\end{document}

%% file: math_commands.tex
\usepackage{amsmath,amsfonts,bm}

\def\eqref#1{equation~\ref{#1}}

\def\1{\bm{1}}

\DeclareMathAlphabet{\mathsfit}{\encodingdefault}{\sfdefault}{m}{sl}
\SetMathAlphabet{\mathsfit}{bold}{\encodingdefault}{\sfdefault}{bx}{n}

%% file: tables/method_comparison.tex
\begin{table}[t]
    \centering
    \setlength{\abovecaptionskip}{2pt}
    \caption{Comparison of how representative methods organize supervision for long-horizon agents. SCAD uses subtask boundaries to define local teacher contexts and subtask--report prefixes to estimate planning credit across completed rollouts. Planning retains signed terminal credit, while execution receives its positive part together with teacher guidance within each bounded subtask context.}
    \label{tab:method-comparison}
    \begingroup
    \fontsize{8}{10}\selectfont
    \setlength{\tabcolsep}{2.5pt}
    \setlength{\aboverulesep}{1.5pt}
    \setlength{\belowrulesep}{1.5pt}
    \renewcommand{\arraystretch}{1.0}
    \renewcommand{\tabularxcolumn}[1]{m{#1}}
    \begin{tabularx}{\textwidth}{@{}l>{\raggedright\arraybackslash}m{74pt}>{\raggedright\arraybackslash}X@{}}
        \toprule
        \textbf{Method} & \textbf{Organizing structure} & \textbf{How supervision is constructed} \\
        \midrule
        HiPER \citep{peng2026hiper}
        & Subgoals and actions
        & Hierarchical advantages for planning and execution \\
        \midrule
        GiGPO \citep{feng2025groupingroup}
        & Shared states
        & Shared-state action comparisons plus trajectory credit \\
        Tree-GRPO \citep{ji2026tree}
        & Rollout trees
        & Branch sampling and tree groups induce implicit step credit \\
        \midrule
        TCOD \citep{wang2026exploring}
        & Trajectory segments
        & Expands student-controlled spans for token distillation \\
        MemOPD \citep{liu2026memopdonpolicydistillationmemory}
        & Invocation states
        & Reconstructs compact invocation states for RL and OPD \\
        ATOD \citep{tan2026atodannealedturnawareonpolicy}
        & Interaction turns
        & Anneals RL/OPD mixing with turn-weighted distillation \\
        \midrule
        \textbf{SCAD (Ours)}
        & \textbf{Subtask--report\newline prefixes}
        & Subtask boundaries set \textbf{local teacher contexts}; subtask--report prefixes organize \textbf{planning credit}; terminal credit is assigned \textbf{by role} \\
        \bottomrule
    \end{tabularx}
    \endgroup
\end{table}

%% file: sections/methodology.tex
\newcounter{scadproposition}
\newcommand{\scadproposition}[2]{\par\noindent\refstepcounter{scadproposition}\label{#1}\textbf{Proposition \thescadproposition.}\ \textit{#2}\par}
\flushbottom
\newcommand{\scaddisplaygaps}[2]{\setlength{\abovedisplayskip}{#1pt}\setlength{\abovedisplayshortskip}{#1pt}\setlength{\belowdisplayskip}{#2pt}\setlength{\belowdisplayshortskip}{#2pt}}

\begin{scadblock}
\noindent\textbf{Problem Formulation.} We study long-horizon tasks with input $x=(q,v)$, comprising a task instruction $q$ and optional visual input $v$. A policy $\pi_\theta$ with parameters $\theta$ generates action $a_t$ at interaction step $t$, then receives observation $o_t$ from the environment's observation law $\mathcal P$. A trajectory $\tau=(a_t,o_t)_{t=1}^{H_\tau}$ contains $H_\tau$ steps. Its history $h_t=(x,a_{<t},o_{<t})$ includes all preceding actions and observations, and $c_t$ is the rendered model context. Given training pairs $(x,y^\star)$ from dataset $\mathcal D$, a terminal evaluator compares the final result with reference outcome $y^\star$. Its binary reward $R(\tau,y^\star)\in\{0,1\}$ equals one when the completed trajectory succeeds and zero otherwise:
\scaddisplaygaps{5.3}{5.7}
\begin{equation}
\label{eq:problem-formulation}
P_\theta(\tau\mid x)
=\prod_{t=1}^{H_\tau}\pi_\theta(a_t\mid c_t)\,\mathcal{P}(o_t\mid h_t,a_t),\qquad
J_{\mathrm{task}}(\theta)
=\mathbb{E}_{(x,y^\star)\sim\mathcal{D},\,\tau\sim P_\theta(\cdot\mid x)}
\big[R(\tau,y^\star)\big].
\end{equation}
Here $P_\theta$ is the trajectory distribution and $J_{\mathrm{task}}$ the task objective; $\mathbb E$ averages over training inputs and interactions. The product separates policy-controlled actions from environment responses, whose joint sequence determines terminal success. A frozen teacher $\pi_T$ supplies intermediate feedback; terminal success remains the objective because local behavior alone does not determine the outcome.
\end{scadblock}

\begin{scadblock}
\subsection{Subtask-Local On-Policy Distillation}
\label{sec:local-distillation}
SCAD uses a shared policy for planning and bounded subtask execution. Each subtask defines a local execution context for on-policy teacher supervision.

\noindent\textbf{Planning and Bounded Execution.} Planning reads the task input and accumulated planning decisions and execution reports. It delegates a self-contained subtask via \texttt{branch} or submits a final output via \texttt{finish}, without calling environment tools. Execution handles the assignment with environment tools and sends a report back via \texttt{return}; it cannot call \texttt{branch} or \texttt{finish}. With environment tools $\mathcal U_{\mathrm{env}}$, the allowed tool sets are $\mathcal U_P=\{\texttt{branch},\texttt{finish}\}$ for planning and $\mathcal U_E=\mathcal U_{\mathrm{env}}\cup\{\texttt{return}\}$ for execution. Let $i$ index complete rollouts and $k$ their subtasks, with instruction $d_{ik}$, report $r_{ik}$, and prior planning history $h^P_{i,<k}$. The corresponding model contexts are
\scaddisplaygaps{6.6}{6.6}
\begin{equation}
\label{eq:local-context}
c^P_{ik}=\mathcal C_P(x,h^P_{i,<k}),\qquad
c^E_{ikt}=\mathcal C_E(x,d_{ik},u_{ik,<t}).
\end{equation}
Superscripts $P,E$ denote planning and execution; $\mathcal C_P,\mathcal C_E$ render contexts with their system prompts, and $u_{ik,<t}$ contains local interactions before token $t$. Each execution starts from $(x,d_{ik})$, excluding generated planning history and earlier execution traces. After it returns, planning appends the report to its history and decides whether to delegate again or finish. We bound local transcript growth by token budget $B$, reserving space for the report. With $|\cdot|$ measuring token length and $O(\cdot)$ bounding growth up to constant factors, token $y_{ikt}$ is sampled from frozen rollout policy $\pi_{\mathrm{old}}$:
\scaddisplaygaps{7.3}{6.1}
\begin{equation}
\label{eq:execution-budget}
|c^E_{ikt}|=O\!\left(|x|+|d_{ik}|+B\right),\qquad
y_{ikt}\sim\pi_{\mathrm{old}}(\cdot\mid c^E_{ikt}).
\end{equation}
The bound includes the fixed task input and subtask instruction; $B$ limits the added local transcript rather than the full task history. Execution reports results and remaining uncertainties when its subtask ends or budget is exhausted. Only this report enters planning history. We instantiate $\mathcal U_{\mathrm{env}}$ with search, page opening, and image search; other tasks may use different tools.
\end{scadblock}

\scadpagebreak
\begin{figure}[t]
    \centering
    \includegraphics[width=\linewidth,height=0.5625\linewidth]{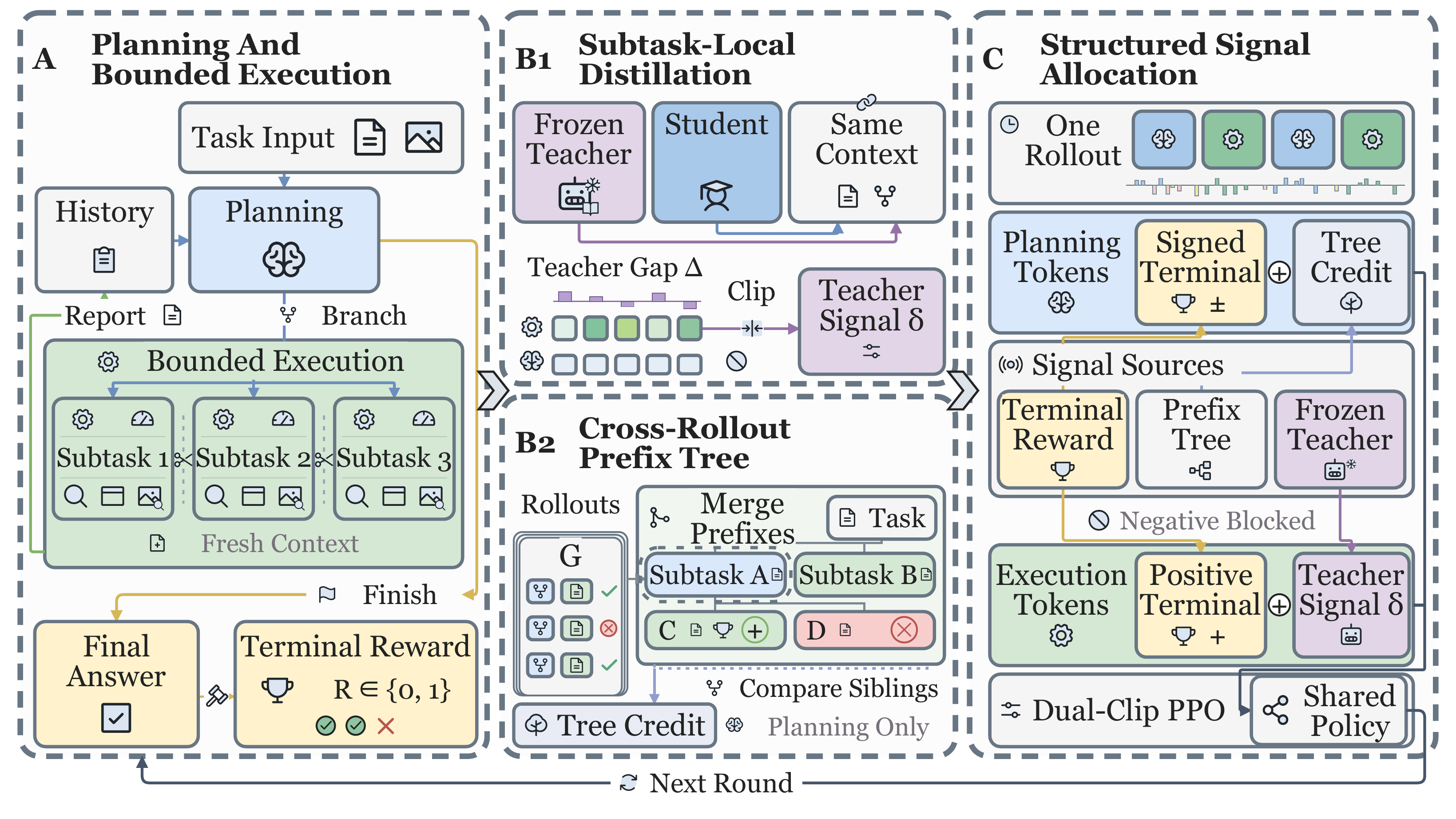}
    \caption[Training framework of SCAD.]{SCAD training framework with local distillation and subtask prefix-tree credit. Planning uses full terminal credit; execution uses positive terminal credit and teacher guidance.}
    \label{fig:scad-method}
\end{figure}

\begin{scadblock}
\noindent\textbf{Teacher Guidance in Matched Contexts.} Following on-policy distillation \citep{agarwal2024onpolicy,gu2024minillm}, the teacher scores student-sampled execution tokens in the same local context as the student; planning tokens receive no teacher supervision. For token $y$ and context $c$, the log-probability gap $\Delta(y,c)$ measures relative teacher preference. Its expectation is negative reverse KL, where $D_{\mathrm{KL}}$ denotes Kullback--Leibler divergence. A positive cap $b$ bounds the signed signal $\delta_{ikt}$ via $\operatorname{clip}$:
\scaddisplaygaps{7.2}{5.9}
\begin{equation}
\label{eq:local-opd}
\begin{aligned}
\Delta(y,c)&=\log\frac{\pi_T(y\mid c)}{\pi_{\mathrm{old}}(y\mid c)},\qquad
\delta_{ikt}=\operatorname{clip}\!\left(\Delta(y_{ikt},c^E_{ikt}),-b,b\right),\\
\mathbb E_{y\sim\pi_{\mathrm{old}}(\cdot\mid c)}[\Delta(y,c)]
&=-D_{\mathrm{KL}}\!\left(\pi_{\mathrm{old}}(\cdot\mid c)\,\|\,\pi_T(\cdot\mid c)\right).
\end{aligned}
\end{equation}
Teacher feedback uses local evidence; collected subtasks can be scored in parallel.
\scadproposition{prop:local-kl}{At fixed local contexts and the rollout policy, bounded policy scores give a reverse-KL gradient approximation whose bias is controlled by clipped log-ratio tails (Appendix~\ref{app:proof-local-kl}).}
\end{scadblock}

\begin{scadblock}
\subsection{Cross-Rollout Planning Credit}
\label{sec:planning-credit}
We build subtask--report prefix trees from $G$ completed rollouts per input, comparing terminal returns under shared histories to estimate credit for subtask choices and stopping decisions.

\noindent\textbf{Prefix Construction and Action Values.} Subtask texts are deduplicated and greedily clustered by embedding cosine similarity. Reports undergo typed normalization and semantic matching, with long reports kept intact. Let $z_{ik}$ be the subtask cluster and $\kappa_{z_{ik}}$ its report canonicalizer. Node $p_{ik}$ follows subtask $k$: $\operatorname{root}$ creates the input's root, and $\operatorname{node}$ identifies a child by its parent, subtask, and report key, keeping distinct canonical histories separate. A \texttt{STOP} branch records the decision to finish:
\scaddisplaygaps{5.6}{6.3}
\begin{equation}
\label{eq:prefix-values}
p_{i0}=\operatorname{root}(x),\qquad
p_{ik}=\operatorname{node}\!\left(p_{i,k-1},z_{ik},\kappa_{z_{ik}}(r_{ik})\right).
\end{equation}
At parent $p$, $\mathcal A(p)$ contains observed subtask clusters and \texttt{STOP}; $\mathcal I(p,a)$ collects distinct rollouts choosing $a$, pooling report variants. Given return $R_i=R(\tau_i,y^\star)$ for rollout $\tau_i$, define visit count $n_{pa}$, Monte Carlo action value $Q(p,a)$, and parent visits $N_p$:
\scaddisplaygaps{7.0}{3.9}
\begin{equation}
\label{eq:action-values}
n_{pa}=|\mathcal I(p,a)|,\qquad
Q(p,a)=\frac{1}{n_{pa}}\sum_{i\in\mathcal I(p,a)}R_i,\qquad
N_p=\sum_{a\in\mathcal A(p)}n_{pa},
\end{equation}
Here $|\mathcal I(p,a)|$ is set cardinality. First-visit counting prevents duplicated training sequences from multiplying a rollout's contribution; the tree and decision spans remain fixed during optimization.
\end{scadblock}

\scadpagebreak
\begin{scadblock}
\noindent\textbf{Sibling Comparisons with Prior Shrinkage.} Let $\mu_x$ and $\sigma_x$ denote the mean and sample standard deviation of the $G$ terminal returns. At each parent prefix, $\bar Q_p$ is the mean of observed action values weighted by their visit counts. We keep these action values unchanged and shrink their mean toward the problem-level mean. The positive prior count $\nu$ controls this shrinkage in baseline $V(p)$; its influence decreases as more rollouts visit the parent, giving observed local returns more weight:
\scaddisplaygaps{7.8}{5.4}
\begin{equation}
\label{eq:tree-credit}
V(p)=\frac{N_p\bar Q_p+\nu\mu_x}{N_p+\nu}
=\bar Q_p+\frac{\nu}{N_p+\nu}(\mu_x-\bar Q_p).
\end{equation}
The local scale $s_p$ is the sample standard deviation of action values repeated by visit count, so frequent choices contribute proportionally. The variance below applies to parents with at least two visits; otherwise the local scale is zero. The effective scale $\widetilde\sigma_p$ uses the local scale if it exceeds the positive tolerance $\epsilon$, otherwise the problem-level standard deviation if above that tolerance, then one. The resulting tree advantage $A^{\mathrm{tree}}$ uses the shared cap $b$ to limit the magnitude of planning credit:
\scaddisplaygaps{7.8}{8.4}
\begin{equation}
\label{eq:tree-scale}
s_p^2=\frac{\sum_{a\in\mathcal A(p)}n_{pa}(Q(p,a)-\bar Q_p)^2}{N_p-1},\qquad
A^{\mathrm{tree}}(p,a)=\operatorname{clip}\!\left(\frac{Q(p,a)-V(p)}{\max(\widetilde\sigma_p,\epsilon)},-b,b\right).
\end{equation}
The same baseline and scale apply to all observed choices under a parent. With one observed choice, single-candidate credit reflects prior shrinkage; with multiple choices, multi-candidate credit also compares actions. Both use Eq.~\ref{eq:tree-scale} only on planning spans; reports match histories without serving as verified subtask outcomes or execution rewards; terminal success supplies the return signal.
\scadproposition{prop:tree-variance}{Before scaling and clipping, tree credit removes within-choice return variance; prior shrinkage shifts its mean without changing sibling contrasts or variance (Appendix~\ref{app:proof-tree-variance}).}
\end{scadblock}

\begin{scadblock}
\subsection{Structured Signal Allocation and Policy Optimization}
\label{sec:signal-allocation}
Planning retains signed terminal and tree credit, while execution combines positive terminal credit with signed local teacher feedback through structured signal allocation across the shared policy.

\noindent\textbf{Terminal Credit and Teacher Guidance.} In rollout $i$, $\mathcal P_i$ and $\mathcal E_i$ index planning and execution tokens. The GRPO advantage $A_i^{\mathrm{grp}}$ is the terminal return minus the group mean, divided by the group standard deviation plus the numerical tolerance \citep{shao2024deepseekmathpushinglimitsmathematical}. Each rollout enters the group once. For token $t$, $B_{it}$ sums tree credit over planning decisions whose spans contain it, including \texttt{STOP}; $\delta_{it}$ is local teacher feedback indexed within the full rollout. The token advantage $A_{it}$ combines these signals with unit weights according to whether the token belongs to planning or execution:
\scaddisplaygaps{8.4}{6.2}
\begin{equation}
\label{eq:structured-advantage}
A_{it}=A_i^{\mathrm{grp}}+B_{it}\quad(t\in\mathcal P_i),\qquad
A_{it}=\max(A_i^{\mathrm{grp}},0)+\delta_{it}\quad(t\in\mathcal E_i).
\end{equation}
Tree credit is zero outside recorded planning decisions. Negative teacher gaps can counteract success reinforcement; when all group outcomes agree, terminal credit vanishes but distillation remains.
\end{scadblock}

\begin{scadblock}
\noindent\textbf{Token-Averaged Policy Update.} For generated token $y_{it}$ in stored context $c_{it}$, $\rho_{it}$ is the current-to-rollout policy probability ratio. The binary mask $m_{it}$ selects valid generated tokens, excluding prompts, observations, padding, and overlong samples; $M$ is their total count. The indicator $\mathbf1[\cdot]$ activates the negative-advantage correction. The PPO surrogate $g_{it}$ uses clip width $\epsilon_c$ \citep{schulman2017proximalpolicyoptimizationalgorithms}; dual clipping with factor $c_d$ greater than one bounds negative-advantage updates \citep{ye2020masteringcomplexcontrolmoba}. We maximize the token-averaged objective $J_{\mathrm{SCAD}}$, keeping rollout probabilities, teacher scores, and all advantages fixed throughout optimization of the shared policy on the collected batch:
\scaddisplaygaps{6.5}{4.2}
\begin{equation}
\label{eq:policy-update}
\begin{aligned}
\rho_{it}&=\frac{\pi_\theta(y_{it}\mid c_{it})}{\pi_{\mathrm{old}}(y_{it}\mid c_{it})},\qquad
g_{it}=\min\!\left(\rho_{it}A_{it},\operatorname{clip}(\rho_{it},1-\epsilon_c,1+\epsilon_c)A_{it}\right),\\
J_{\mathrm{SCAD}}(\theta)
&=\frac{1}{M}\sum_{i,t}m_{it}\left[g_{it}
+\mathbf{1}[A_{it}<0]\max(c_dA_{it}-g_{it},0)\right].
\end{aligned}
\end{equation}
Each update grades rollouts, builds the prefix tree, and scores execution tokens before optimizing the shared policy. Teacher and embedding models remain frozen. All supervision enters through allocated advantages, without separate reference-KL or entropy terms for planning or execution.
\scadproposition{prop:allocation-gradient}{Under dual clipping, execution allocation weakly increases each token's direct log-probability derivative over uniform fusion and equals teacher-only updates on failures (Appendix~\ref{app:proof-allocation-gradient}).}
\end{scadblock}

%% file: sections/experiments.tex
\begingroup
\flushbottom
\setlength{\parskip}{3pt plus 2pt minus 0.5pt}
\setlength{\intextsep}{5pt}
\setlength{\abovecaptionskip}{4pt}
\setlength{\belowcaptionskip}{0pt}
\setlength{\emergencystretch}{0.8em}
\definecolor{expReference}{RGB}{242,245,250}
\definecolor{expTeacher}{RGB}{224,214,245}
\definecolor{expRL}{RGB}{239,246,253}
\definecolor{expDistill}{RGB}{255,248,238}
\definecolor{expOurs}{RGB}{235,248,237}
\newcommand{\expstd}[2]{\mbox{#1\kern0.25pt\raisebox{-1.3pt}[0pt][0pt]{\color{black!75}\fontsize{4.6}{4.6}\selectfont\textpm#2}}}
\newcommand{\expbeststd}[2]{\expstd{\textbf{#1}}{#2}}
\makeatletter
\def\subsection{\@startsection{subsection}{2}{\z@}{-6pt}{2pt}{\normalsize\sc\raggedright}}
\newenvironment{experimenttable}{\par\addvspace{4pt}\noindent\begin{minipage}{\textwidth}\def\@captype{table}}{\end{minipage}\par}
\newenvironment{experimentfigure}{\par\addvspace{4pt}\nointerlineskip\vbox\bgroup\hsize=\textwidth\@parboxrestore\def\@captype{figure}\centering\setlength{\parskip}{0pt}}{\par\egroup\par\addvspace{4pt}}
\makeatother
\newcommand{\exppanelcaption}[2]{\par\vspace{3pt}\noindent\parbox{\linewidth}{\small\microtypesetup{expansion=true}\parfillskip=0pt Figure~\thefigure(#1): #2\par}\par}
\newcommand{\expintro}[2][0pt]{\par\begingroup\microtypesetup{expansion=true}\parfillskip=#1\relax\noindent#2\par\endgroup}
\newcommand{\expdataset}[2]{{\fontsize{6}{8}\rmfamily\bfseries\selectfont\makebox[\linewidth][c]{\if\relax\detokenize{#1}\relax
        \raisebox{4.2pt}{\strut #2}\else
        \raisebox{-0.8pt}{\shortstack[c]{\strut #1\\[-1pt]\strut #2}}\fi}}}
\input{tables/text_results}
\input{tables/multimodal_results}

\section{Experiments}
\label{sec:experiments}
We address six research questions. \textbf{RQ1:} Does SCAD improve text and multimodal task performance over existing training baselines? \textbf{RQ2:} How much does each training component contribute? \textbf{RQ3:} Does local distillation sustain useful teacher supervision at greater interaction depths? \textbf{RQ4:} Does prefix-tree credit better distinguish planning decisions under shared histories? \textbf{RQ5:} Does structured signal allocation reconcile local execution quality with global success? \textbf{RQ6:} How efficiently does SCAD use teacher supervision and training computation?

\subsection{Experimental Setup}
\label{sec:experimental-setup}
\noindent\textbf{Datasets and Evaluation Metrics.} ID datasets occur in training; OOD datasets do not (Appendix~\ref{app:resource-sources}). Test sets contain 814 text (525 ID, 289 OOD) and 510 multimodal (395 ID, 115 OOD) questions. Final-answer accuracy (\%) uses greedy decoding and Gemini-3.7-Flash as judge. Avg. columns and figure accuracies equally weight all 11 text or 8 multimodal datasets. Checkpoints use validation macro accuracy, including Bamboogle, PopQA, and FVQA.

\noindent\textbf{Baselines and Ablations.} Tables~\ref{tab:text-results}--\ref{tab:multimodal-results} include Vanilla, teacher, RL, supervised, distillation, and hybrid baselines; Vanilla directly generates answers without planning or retrieval. RQ1 examines SCAD training dynamics; RQ2 adds local OPD, tree credit, and signal allocation within a fixed planning/execution architecture. RQ3--RQ5 analyze the three designs of Sections~\ref{sec:local-distillation}--\ref{sec:signal-allocation}. RQ1--RQ2 cover both modalities; RQ3--RQ6 use text diagnostics. The cumulative ablation measures accuracy changes between successive configurations as components are added in a fixed order.

\noindent\textbf{Implementation Details.} Training uses 400 text and 347 multimodal examples. SCAD uses 60 steps, eight prompts with eight rollouts each, and AdamW at $10^{-6}$. Each Worker has a local transcript growth budget of 16,384 tokens. All three signal weights are one, with teacher/tree signals clipped at five. Text training uses four RTX PRO 6000 GPUs, selecting each checkpoint by validation macro accuracy. Subtask and report thresholds are 0.83 and 0.90, with the prior count $\nu=2$.

\scadpagebreak
\setlength{\parskip}{3pt plus 4pt minus 0.5pt}
\makeatletter
\let\exporiginalmakecaption\@makecaption
\long\def\@makecaption#1#2{\exporiginalmakecaption{#1}{\parfillskip=0pt #2}}
\makeatother
\newenvironment{expsubcaptions}[1][396]{\par\nointerlineskip\vspace{1pt}\noindent\setlength{\unitlength}{1bp}\small\begin{picture}(#1,6.5)}{\end{picture}\par\vspace{-1pt}}
\newcommand{\expsub}[2]{\put(#1,0){\makebox[0pt]{#2}}}
\newcommand{\expblockcaption}[1]{\par\vspace{3pt}\noindent\parbox{\linewidth}{\small\microtypesetup{expansion=true}\parfillskip=0pt Figure~\thefigure: #1\par}\par}
\subsection{Main Results (RQ1)}
\label{sec:rq1}
\expintro{Tables~\ref{tab:text-results}--\ref{tab:multimodal-results} compare task accuracy across methods, while Figure~\ref{fig:rq1} pairs SCAD's accuracy and teacher--student divergence. The four panels jointly cover text and multimodal training dynamics.}

\begin{experimentfigure}
\refstepcounter{figure}\label{fig:rq1}
\begin{minipage}[t]{0.475\textwidth}
\vspace{0pt}\microtypesetup{expansion=true}
\setlength{\parskip}{3pt}
\noindent\textbf{Task performance.} SCAD reaches 46.10\% text and 33.13\% multimodal overall macro accuracy. The strongest trained baselines are ATOD at 41.62\% and HyperEyes at 28.94\%, yielding gains of 4.48 and 4.19 points. Relative to Vanilla at 17.64/11.61\%, the gains are 28.46/21.52 points. SCAD leads the trained baselines across individual datasets, and the multimodal student surpasses its teacher by 2.23 points in overall accuracy.

\noindent\textbf{Training dynamics.} Overall accuracy rises from 38.91 to 46.10\% for text and from 16.28 to 33.13\% for multimodal tasks. Reverse KL decreases from 0.468 to 0.253 nats/token for text and from 0.356 to 0.229 for multimodal execution. This diagnostic compares student and teacher distributions on fixed execution prefixes. Lower KL reflects closer alignment; accuracy separately measures task performance. Training initialization uses the SCAD scaffold and is tracked separately from the Vanilla table entries. SCAD starts lower but finishes above both baselines within a 60-step training run.\looseness=-1
\end{minipage}\hfill
\begin{minipage}[t]{198bp}
\vspace{0pt}\centering
\includegraphics{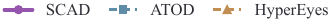}\par\vspace{3pt}
\includegraphics{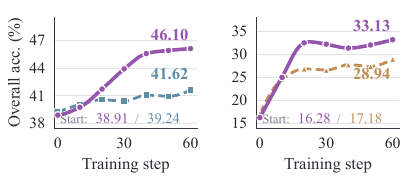}
\begin{expsubcaptions}[198]
\expsub{60.35}{(a) Text accuracy}\expsub{153.5}{(b) Multimodal accuracy}
\end{expsubcaptions}
\vspace{9pt}
\includegraphics{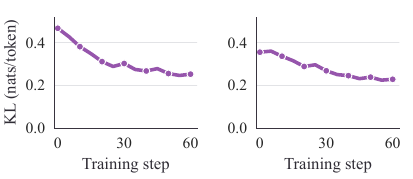}
\begin{expsubcaptions}[198]
\expsub{60.35}{(c) Text KL}\expsub{153.5}{(d) Multimodal KL}
\end{expsubcaptions}
\vspace{2.37pt}\expblockcaption{Training dynamics over the 60-step run. Panels (a,b) compare SCAD's accuracy with the strongest trained baselines, ATOD and HyperEyes, labeling both step-0 values as Start; (c,d) show SCAD's reverse KL.}
\end{minipage}
\end{experimentfigure}

\subsection{Ablation Study (RQ2)}
\label{sec:rq2}
\expintro{Figure~\ref{fig:rq2} adds the three training components within a fixed local planning/execution architecture. The two waterfalls compare their cumulative contributions to text and multimodal overall accuracy.}

\begin{experimentfigure}
\includegraphics[width=\linewidth]{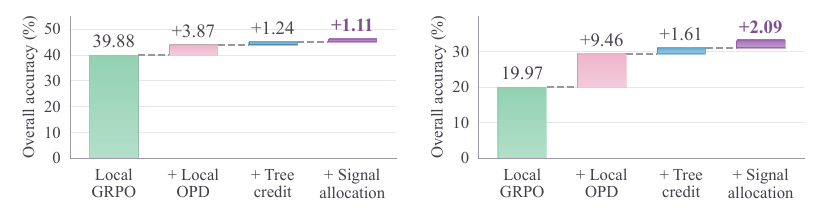}
\begin{expsubcaptions}
\expsub{111.9}{(a) Text}\expsub{307.9}{(b) Multimodal}
\end{expsubcaptions}
\caption{Cumulative overall accuracy gains as local OPD, tree credit, and structured signal allocation are added in turn to local GRPO for (a) text and (b) multimodal tasks, averaged over 3 seeds.}
\label{fig:rq2}
\end{experimentfigure}
\noindent\textbf{Component contributions.} Local OPD adds 3.87/9.46 points to the text/multimodal starting accuracies of 39.88/19.97\%. Tree credit adds 1.24/1.61 and structured allocation adds 1.11/2.09, reaching 46.10/33.13\%. Local OPD accounts for 62.2/71.9\% of the total gain in this sequence. Tree credit and allocation jointly add 2.35/3.70 points, providing complementary gains once local teacher supervision is available and consistently improving accuracy across both modalities.

\noindent\textbf{Complementary signals.} Tree credit refines planning supervision, while allocation controls execution updates, adding 2.09 points for multimodal tasks versus 1.11 for text. Total gains are 6.22/13.16 points. Components enter in the order the method composes them: local OPD supplies execution supervision first, and tree credit and allocation then build on it. The gains combine execution-level guidance with outcome-aware planning updates within the same interaction architecture. RQ3--RQ5 below examine the mechanism behind each of these components.

\scadpagebreak
\subsection{Analysis of Subtask-Local Distillation (RQ3)}
\label{sec:rq3}
\expintro{Panels (a,b) in Figure~\ref{fig:rq3} examine teacher supervision on 200 long-horizon questions from the text benchmarks. We use an untrained Qwen3-4B student and trajectories reaching interaction 24.}

\begin{experimentfigure}
\includegraphics[width=\linewidth]{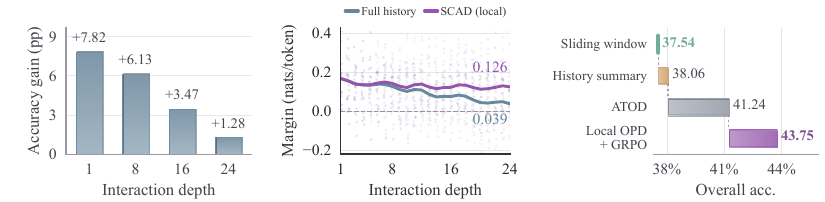}
\begin{expsubcaptions}
\expsub{77.2}{(a) Teacher continuation}\expsub{204.0}{(b) Teacher preference}\expsub{325.9}{(c) Context variants}
\end{expsubcaptions}
\caption{(a) Teacher--student continuation gain (pp). (b) Teacher score margin (nats/token). (c) Overall text ID+OOD accuracy (\%), averaged over 3 seeds; floating bars link adjacent settings.}
\label{fig:rq3}
\end{experimentfigure}
\noindent\textbf{Teacher continuation and preference.} In (a), teacher and student continue from identical prefixes at depths 1/8/16/24 with equal budgets. Bars show the difference in final-answer accuracy, averaged equally across datasets: 7.82 to 1.28 points. Teacher continuation is diagnostic only. In (b), the teacher scores paired correct/incorrect candidates at the same state under full-history and local contexts. The margin is correct minus incorrect mean token log-probability; positive values favor correct execution. Dots show candidate pairs and lines show depthwise means. Local/full-history margins at depth 24 are 0.126/0.039 nats/token, showing clearer preferences with local contexts.

\noindent\textbf{Accuracy after training.} Sliding window and history summary (OPD) reach 37.54\% and 38.06\%. Within the Planner--Worker framework, ATOD reaches 41.24\% using full-history teacher supervision without subtask resets. Local OPD + GRPO reaches 43.75\%, matching RQ2's second stage and exceeding ATOD by 2.51 points. This variant uses ordinary Planner GRPO and signed Worker GRPO plus local OPD. Scores equally weight seven ID and four OOD text datasets.

\subsection{Analysis of Cross-Rollout Planning Credit (RQ4)}
\label{sec:rq4}
\expintro{Figure~\ref{fig:rq4} compares prefix-tree credit and uniform planning-advantage scaling. We isolate single- and multi-candidate credit while keeping the execution-side supervision fixed across all five settings.}

\begin{experimentfigure}
\refstepcounter{figure}\label{fig:rq4}
\begin{minipage}[t]{0.48\textwidth}
\vspace{0pt}\centering
\includegraphics[width=\linewidth]{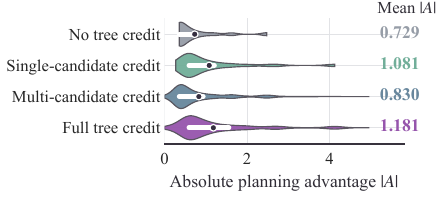}
\exppanelcaption{a}{Absolute planning advantages $|A|$; dots show means and white bars the interquartile range.}
\vspace{8pt}
\includegraphics[width=\linewidth]{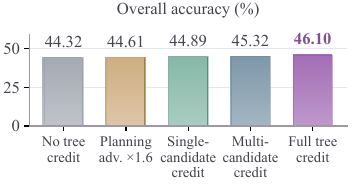}
\exppanelcaption{b}{Overall text ID+OOD accuracy across five planning-credit variants, averaged over 3 seeds.}
\end{minipage}\hfill
\begin{minipage}[t]{0.495\textwidth}
\vspace{0pt}\microtypesetup{expansion=true}
\setlength{\parskip}{5.35pt}
\noindent\textbf{Credit settings.} No tree disables the tree term. Single/multi-candidate credit activates it at prefixes with one/at least two distinct actions; full tree covers both (Eq.~\ref{eq:tree-scale}). All retain terminal advantages. The $\times1.6$ control scales normalized no-tree Planner advantages before the policy loss; Workers retain positive GRPO and local OPD.

\noindent\textbf{Multi-candidate coverage.} These groups account for 23.51\% of planning-decision visits and occur at least once in 72.66\% of agent trajectories.

\noindent\textbf{Credit and performance.} On the same 1,146 reconstructed decisions, the four rules yield mean $|A|$ of 0.729/1.081/0.830/1.181. Uniform scaling gives 1.167 and raises accuracy from 44.32\% to 44.61\%. Full tree credit reaches 46.10\%, 1.49 points above scaling, supporting prefix-specific planning credit. The comparison holds Worker supervision fixed and changes only Planner credit, linking distinctions under shared histories to final task accuracy under the common training budget and evaluation protocol across all five variants.
\end{minipage}
\end{experimentfigure}

\scadpagebreak
\setlength{\parskip}{2pt plus 2pt minus 0.5pt}
\subsection{Analysis of Structured Signal Allocation (RQ5)}
\label{sec:rq5}
\expintro{Panels (a,b) in Figure~\ref{fig:rq5} use 200 text ID+OOD questions and an untrained Qwen3-4B student. Subtask labels reflect semantic agreement with reference answers generated using Gemini-3.7-Flash.}

\begin{experimentfigure}
\includegraphics[width=\linewidth]{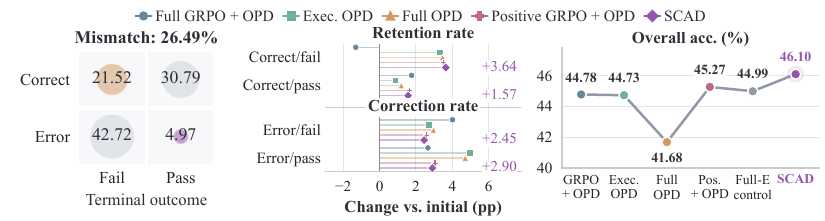}
\begin{expsubcaptions}
\expsub{70.3}{(a) Label agreement}\expsub{203.1}{(b) Retention and correction}\expsub{331.45}{(c) Allocation variants}
\end{expsubcaptions}
\caption{(a) Joint subtask/terminal proportions (\%). (b) Retention/correction changes at shared states. (c) Overall text ID+OOD accuracy under all six allocation settings, averaged over 3 seeds.}
\label{fig:rq5}
\end{experimentfigure}
\noindent\textbf{Allocation settings.} All six variants share our Planner--Worker framework. Full/Positive GRPO + OPD apply signed/positive terminal credit plus OPD to both modules. Execution-only OPD, Full-E control, and SCAD share signed planning credit without OPD; their Workers use OPD alone, signed GRPO + OPD, and positive GRPO + OPD, respectively. All GRPO variants retain signed planning tree credit; Full OPD distills both modules without GRPO or tree credit.

\noindent\textbf{Local behavior and overall accuracy.} Subtask and terminal labels disagree on 26.49\% of subtasks. Positive GRPO + OPD and SCAD improve correct/failed retention by 3.51/3.64 points. With planning fixed, Full-E control reaches 44.99\% versus SCAD's 46.10\%; positive-only execution credit adds 1.11 points, matching RQ2's allocation increment. This supports role-specific supervision.

\subsection{Token and Computational Efficiency (RQ6)}
\label{sec:rq6}
\expintro[0pt plus 1fil]{All three methods in Figure~\ref{fig:rq6}(a) share the same Planner--Worker framework for subtask planning and execution. Local contexts allow parallel teacher scoring, reducing stragglers.}

\begin{experimentfigure}
\includegraphics[width=\linewidth]{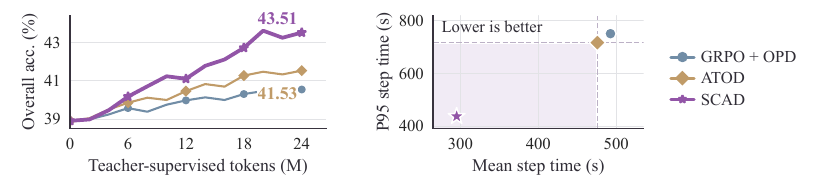}
\begin{expsubcaptions}
\expsub{95.05}{(a) Supervision efficiency}\expsub{258.4}{(b) Training step time}
\end{expsubcaptions}
\caption{Means over 3 runs. (a) Accuracy versus cumulative teacher-supervised tokens. (b) Mean/P95 step time excluding Gemini-3.7-Flash judging; dashed lines and shading denote ATOD.}
\label{fig:rq6}
\end{experimentfigure}
\noindent\textbf{Teacher-supervision efficiency.} At 24 million teacher-supervised tokens, SCAD reaches 43.51\% overall accuracy versus 41.53\% for ATOD, a 1.98-point gap. The budget counts only response tokens scored by the teacher for OPD, not all student rollout tokens. Curves thus compare accuracy at equal amounts of teacher guidance, which SCAD allocates to local execution.

\noindent\textbf{Training time.} Excluding Gemini-3.7-Flash judging, 60 steps take 4.92 hours for SCAD, 7.93 for ATOD, and 8.21 for GRPO + OPD. SCAD uses 37.96\% less time than ATOD.

\makeatletter
\def\section{\@startsection{section}{1}{\z@}{-6pt}{3pt}{\large\sc\raggedright}}
\makeatother
\section{Conclusion}
\label{sec:conclusion}
We presented SCAD, which combines subtask-local distillation, prefix-tree planning credit, and structured signal allocation for long-horizon agents. Planning receives full terminal credit, while execution combines positive terminal credit with local teacher feedback. Across all benchmarks, macro accuracy improves over the strongest trained baselines by 4.48 points on text tasks and 4.19 on multimodal tasks. SCAD uses no process rewards, learned critic, or extra counterfactual branch sampling. Future work will explore adaptive allocation of teacher supervision across subtasks, guided by subtask difficulty and the expected learning benefit within a fixed teacher computation budget.
\endgroup

%% file: tables/text_results.tex
\begin{experimenttable}
    \centering
    \setlength{\abovecaptionskip}{0pt}
    \setlength{\belowcaptionskip}{4pt}
    \caption{Text accuracy (\%); Qwen3-4B(student) / Qwen3-4B-Instruct-2507 (teacher). Mean$\pm$std over 3 seeds. Avg.: equal-weight mean over 11 ID+OOD sets; \textdagger: OOD. Bold: best student result.}
    \label{tab:text-results}
    \begingroup
    \fontsize{7}{7.7}\selectfont
    \setlength{\tabcolsep}{0.1pt}
    \setlength{\aboverulesep}{1.0pt}
    \setlength{\belowrulesep}{1.0pt}
    \renewcommand{\arraystretch}{1.28}
    \begin{tabularx}{\textwidth}{@{}l*{12}{>{\centering\arraybackslash}X}@{}}
        \toprule
        \raisebox{4.2pt}{\textbf{Method}} & \expdataset{Browse}{Comp-Plus} & \expdataset{WebWalker}{QA} & \expdataset{}{Hotpot} & \expdataset{}{2Wiki} & \expdataset{}{MuS} & \expdataset{}{NQ} & \expdataset{}{Trivia} & \expdataset{}{Bamb.\textsuperscript{\textdagger}} & \expdataset{}{PopQA\textsuperscript{\textdagger}} & \expdataset{}{GAIA\textsuperscript{\textdagger}} & \expdataset{}{xbench\textsuperscript{\textdagger}} & \expdataset{}{Avg.} \\
        \midrule
        \rowcolor{expTeacher}
        Teacher & \expstd{23.61}{0.5} & \expstd{36.67}{3.1} & \expstd{60.51}{0.9} & \expstd{52.31}{1.5} & \expstd{11.28}{0.9} & \expstd{68.72}{1.8} & \expstd{85.13}{0.9} & \expstd{56.92}{0.0} & \expstd{56.92}{0.0} & \expstd{25.42}{0.0} & \expstd{23.67}{0.6} & \expstd{45.56}{0.6} \\
        \midrule
        \rowcolor{expReference}
        Vanilla & \expstd{0.00}{0.0} & \expstd{2.92}{0.7} & \expstd{31.28}{0.9} & \expstd{21.54}{0.0} & \expstd{11.79}{0.9} & \expstd{28.21}{0.9} & \expstd{46.67}{1.8} & \expstd{24.10}{0.9} & \expstd{21.03}{0.9} & \expstd{2.82}{1.0} & \expstd{3.67}{0.6} & \expstd{17.64}{0.5} \\
        \rowcolor{expRL}
        GRPO & \expstd{19.44}{4.6} & \expstd{28.75}{1.3} & \expstd{40.00}{1.5} & \expstd{46.15}{1.5} & \expstd{10.77}{1.5} & \expstd{63.59}{1.8} & \expstd{75.90}{1.8} & \expstd{32.82}{5.4} & \expstd{42.05}{6.2} & \expstd{15.25}{0.0} & \expstd{25.00}{3.6} & \expstd{36.34}{1.4} \\
        \rowcolor{expRL}
        FoldGRPO & \expstd{17.22}{0.5} & \expstd{34.58}{5.1} & \expstd{52.31}{5.5} & \expstd{52.31}{1.5} & \expstd{13.33}{0.9} & \expstd{63.08}{0.0} & \expstd{71.28}{1.8} & \expstd{41.54}{1.5} & \expstd{55.38}{1.5} & \expstd{13.56}{6.1} & \expstd{21.00}{1.0} & \expstd{39.60}{0.9} \\
        \rowcolor{expRL}
        Tree-GRPO & \expstd{18.33}{2.2} & \expstd{35.83}{0.7} & \expstd{50.77}{0.0} & \expstd{51.28}{1.8} & \expstd{18.46}{1.5} & \expstd{61.03}{0.9} & \expstd{78.46}{8.1} & \expstd{38.46}{4.1} & \expstd{52.31}{1.5} & \expstd{18.64}{1.7} & \expstd{25.33}{4.5} & \expstd{40.81}{0.3} \\
        \rowcolor{expRL}
        HiPER & \expstd{21.67}{2.5} & \expstd{32.92}{3.1} & \expstd{54.36}{4.7} & \expstd{46.15}{4.1} & \expstd{16.92}{1.5} & \expstd{66.15}{1.5} & \expstd{73.85}{1.5} & \expstd{43.59}{0.9} & \expstd{53.85}{1.5} & \expstd{18.64}{1.7} & \expstd{28.33}{1.2} & \expstd{41.49}{0.5} \\
        \rowcolor{expDistill}
        SFT & \expstd{19.17}{3.8} & \expstd{33.33}{7.5} & \expstd{47.69}{1.5} & \expstd{48.21}{5.4} & \expstd{11.79}{1.8} & \expstd{63.08}{1.5} & \expstd{71.28}{0.9} & \expstd{40.00}{1.5} & \expstd{52.31}{1.5} & \expstd{17.51}{9.3} & \expstd{22.33}{1.2} & \expstd{38.79}{2.2} \\
        \rowcolor{expDistill}
        OPD & \expstd{19.72}{3.4} & \expstd{28.75}{1.3} & \expstd{51.79}{0.9} & \expstd{45.64}{1.8} & \expstd{13.85}{1.5} & \expstd{65.64}{5.4} & \expstd{71.79}{5.4} & \expstd{37.44}{0.9} & \expstd{47.18}{6.2} & \expstd{11.86}{1.7} & \expstd{25.00}{0.0} & \expstd{38.06}{1.0} \\
        \rowcolor{expDistill}
        SDAR & \expstd{15.83}{3.8} & \expstd{35.00}{1.3} & \expstd{44.10}{0.9} & \expstd{54.87}{1.8} & \expstd{8.72}{7.6} & \expstd{58.46}{1.5} & \expstd{82.56}{6.2} & \expstd{35.90}{1.8} & \expstd{56.41}{5.4} & \expstd{15.82}{1.0} & \expstd{26.67}{1.2} & \expstd{39.49}{2.2} \\
        \rowcolor{expDistill}
        ATOD & \expstd{21.39}{3.4} & \expstd{33.75}{3.3} & \expstd{53.85}{1.5} & \expstd{50.77}{0.0} & \expstd{13.85}{1.5} & \expstd{64.10}{0.9} & \expstd{75.90}{1.8} & \expstd{45.13}{8.5} & \expstd{56.92}{1.5} & \expstd{17.51}{8.4} & \expstd{24.67}{0.6} & \expstd{41.62}{1.0} \\
        \rowcolor{expDistill}
        TCOD & \expstd{20.83}{0.0} & \expstd{30.83}{0.7} & \expstd{47.69}{1.5} & \expstd{49.23}{1.5} & \expstd{11.79}{0.9} & \expstd{59.49}{0.9} & \expstd{80.00}{0.0} & \expstd{41.54}{8.1} & \expstd{50.77}{1.5} & \expstd{14.12}{1.0} & \expstd{22.00}{0.0} & \expstd{38.94}{0.7} \\
        \rowcolor{expDistill}
        MemOPD & \expstd{21.67}{3.0} & \expstd{36.25}{1.3} & \expstd{50.26}{0.9} & \expstd{46.67}{1.8} & \expstd{14.87}{1.8} & \expstd{62.56}{0.9} & \expstd{76.92}{5.5} & \expstd{44.10}{0.9} & \expstd{56.92}{1.5} & \expstd{22.60}{1.0} & \expstd{20.67}{3.5} & \expstd{41.23}{0.9} \\
        \midrule
        \rowcolor{expOurs}
        \textbf{SCAD (Ours)} & \expbeststd{23.06}{2.9} & \expbeststd{40.42}{2.6} & \expbeststd{55.38}{1.5} & \expbeststd{56.41}{2.4} & \expbeststd{18.97}{1.8} & \expbeststd{71.28}{3.9} & \expbeststd{84.10}{1.8} & \expbeststd{45.64}{0.9} & \expbeststd{58.97}{6.2} & \expbeststd{23.16}{3.5} & \expbeststd{29.67}{0.6} & \expbeststd{46.10}{1.2} \\
        \bottomrule
    \end{tabularx}
    \endgroup
\end{experimenttable}

%% file: tables/multimodal_results.tex
\begin{experimenttable}
    \centering
    \setlength{\abovecaptionskip}{0pt}
    \setlength{\belowcaptionskip}{4pt}
    \caption{Multimodal accuracy (\%); Qwen3-VL-2B(student) / Qwen3-VL-4B (teacher). Mean$\pm$std over 3 seeds. ID/All Avg.: equal-weight means over five/eight datasets; \textdagger: OOD. Bold: best student.}
    \label{tab:multimodal-results}
    \begingroup
    \fontsize{7}{7.7}\selectfont
    \setlength{\tabcolsep}{0.1pt}
    \setlength{\aboverulesep}{1.0pt}
    \setlength{\belowrulesep}{1.0pt}
    \renewcommand{\arraystretch}{1.18}
    \begin{tabularx}{\textwidth}{@{}l*{10}{>{\centering\arraybackslash}X}@{}}
        \toprule
        \raisebox{4.2pt}{\textbf{Method}} & \expdataset{MMSearch-}{Plus} & \expdataset{}{VDR-Bench} & \expdataset{VisBrowse-}{Bench} & \expdataset{}{MMSearch} & \expdataset{}{SimpleVQA} & \expdataset{ID}{Avg.} & \expdataset{}{FVQA\textsuperscript{\textdagger}} & \expdataset{}{InfoSeek\textsuperscript{\textdagger}} & \expdataset{}{LiveVQA\textsuperscript{\textdagger}} & \expdataset{All}{Avg.} \\
        \midrule
        \rowcolor{expTeacher}
        Teacher & \expstd{16.19}{1.0} & \expstd{16.79}{0.9} & \expstd{5.64}{1.8} & \expstd{40.61}{1.0} & \expstd{45.71}{2.9} & \expstd{24.99}{1.0} & \expstd{56.19}{1.6} & \expstd{37.14}{2.9} & \expstd{28.89}{2.2} & \expstd{30.90}{1.0} \\
        \midrule
        \rowcolor{expReference}
        Vanilla & \expstd{0.95}{0.0} & \expstd{0.74}{0.7} & \expstd{1.54}{1.5} & \expstd{23.64}{1.8} & \expstd{20.00}{2.9} & \expstd{9.37}{1.2} & \expstd{25.71}{2.9} & \expstd{11.43}{0.0} & \expstd{8.89}{0.0} & \expstd{11.61}{0.4} \\
        \rowcolor{expRL}
        GRPO & \expstd{3.49}{2.4} & \expstd{3.46}{1.9} & \expstd{0.51}{0.9} & \expstd{40.00}{1.8} & \expstd{42.86}{0.0} & \expstd{18.06}{0.5} & \expstd{47.62}{1.6} & \expstd{27.62}{1.6} & \expstd{19.26}{2.6} & \expstd{23.10}{0.4} \\
        \rowcolor{expRL}
        MMSearch-R1 & \expstd{9.52}{1.0} & \expstd{12.35}{1.9} & \expstd{5.13}{0.9} & \expstd{40.61}{4.6} & \expstd{46.67}{10.0} & \expstd{22.85}{3.5} & \expstd{56.19}{1.6} & \expstd{32.38}{1.6} & \expstd{22.22}{2.2} & \expstd{28.13}{2.6} \\
        \rowcolor{expRL} VSearcher & \expstd{10.16}{3.3} & \expstd{14.32}{6.0} & \expstd{6.15}{1.5} & \expstd{37.58}{2.1} & \expstd{47.62}{1.6} & \expstd{23.17}{1.5} & \expstd{51.43}{2.9} & \expstd{36.19}{1.6} & \expstd{23.70}{2.6} & \expstd{28.39}{1.7} \\
        \rowcolor{expDistill}
        SFT & \expstd{4.76}{4.4} & \expstd{6.67}{0.7} & \expstd{2.56}{2.4} & \expstd{35.76}{6.4} & \expstd{39.05}{1.6} & \expstd{17.76}{0.9} & \expstd{48.57}{0.0} & \expstd{24.76}{1.6} & \expstd{17.04}{7.8} & \expstd{22.40}{0.7} \\
        \rowcolor{expDistill}
        OPD & \expstd{7.30}{3.3} & \expstd{9.88}{0.4} & \expstd{2.56}{0.9} & \expstd{33.33}{4.6} & \expstd{38.10}{1.6} & \expstd{18.23}{1.7} & \expstd{54.29}{2.9} & \expstd{27.62}{1.6} & \expstd{19.26}{1.3} & \expstd{24.04}{0.8} \\
        \rowcolor{expDistill}
        VOLD & \expstd{10.79}{3.8} & \expstd{10.12}{2.6} & \expstd{3.08}{1.5} & \expstd{36.36}{1.8} & \expstd{49.52}{1.6} & \expstd{21.98}{0.3} & \expstd{50.48}{3.3} & \expstd{33.33}{1.6} & \expstd{26.67}{2.2} & \expstd{27.54}{0.9} \\
        \rowcolor{expDistill}
        HyperEyes & \expstd{11.43}{2.5} & \expstd{12.59}{0.7} & \expstd{5.64}{3.9} & \expstd{36.36}{1.8} & \expstd{49.52}{10.0} & \expstd{23.11}{3.0} & \expstd{53.33}{1.6} & \expstd{35.24}{1.6} & \expstd{27.41}{5.6} & \expstd{28.94}{2.6} \\
        \midrule
        \rowcolor{expOurs}
        \textbf{SCAD (Ours)} & \expbeststd{15.24}{3.8} & \expbeststd{15.56}{0.7} & \expbeststd{8.21}{6.4} & \expbeststd{41.21}{6.4} & \expbeststd{51.43}{2.9} & \expbeststd{26.33}{1.7} & \expbeststd{59.05}{1.6} & \expbeststd{44.76}{7.2} & \expbeststd{29.63}{1.3} & \expbeststd{33.13}{1.5} \\
        \bottomrule
    \end{tabularx}
    \endgroup
\end{experimenttable}

%% file: sections/additional_experimental_results.tex
\begingroup
\raggedbottom
\setlength{\parskip}{4pt plus 1pt minus .5pt}
\setlength{\intextsep}{6pt}
\setlength{\abovecaptionskip}{4pt}
\setlength{\belowcaptionskip}{0pt}
\definecolor{expReference}{RGB}{242,245,250}
\definecolor{expTeacher}{RGB}{224,214,245}
\definecolor{expRL}{RGB}{239,246,253}
\definecolor{expDistill}{RGB}{255,248,238}
\definecolor{expOurs}{RGB}{235,248,237}
\newcommand{\expstd}[2]{\mbox{#1\kern0.25pt\raisebox{-1.3pt}[0pt][0pt]{\color{black!75}\fontsize{4.6}{4.6}\selectfont\textpm#2}}}
\newcommand{\expbeststd}[2]{\expstd{\textbf{#1}}{#2}}
\makeatletter
\newenvironment{experimenttable}{\par\addvspace{4pt}\noindent\begin{minipage}{\textwidth}\def\@captype{table}}{\end{minipage}\par}
\makeatother
\newcommand{\expdataset}[2]{{\fontsize{6}{8}\rmfamily\bfseries\selectfont\makebox[\linewidth][c]{\if\relax\detokenize{#1}\relax
        \raisebox{4.2pt}{\strut #2}\else
        \raisebox{-0.8pt}{\shortstack[c]{\strut #1\\[-1pt]\strut #2}}\fi}}}

\section{Additional Experimental Results}
\label{app:additional-results}
\subsection{Additional Student Experiments}
\label{app:smaller-student}
\input{tables/text_results_1p7b}

\noindent\textbf{Smaller-student setting.} We train Qwen3-1.7B with Qwen3-4B-Instruct-2507 as teacher and the 4B setting's architecture. Training uses 400 examples, 60 updates, eight prompts with eight rollouts each, and a learning rate of $10^{-6}$ on two RTX PRO 6000 GPUs. Checkpoints use macro accuracy on 130 validation examples. Evaluation covers 525 ID and 289 OOD questions; Avg. equally weights all eleven datasets. Gemini-3.7-Flash applies the same final-answer criteria across student sizes. We use greedy decoding and score final answers by semantic agreement with the provided references.

\noindent\textbf{Main comparison.} SCAD reaches 39.58\% overall accuracy, exceeding ATOD's 35.32\% and TCOD's 35.04\% (Table~\ref{tab:text-results-1p7b}). The 4.26-point gain over the strongest training baseline extends the main result to a smaller student. SCAD leads all eleven student columns, with 42.50\% on WebWalkerQA, 47.18\% on 2Wiki, and 36.00\% on xbench, covering navigation, multi-hop reasoning, and search.

\begin{wrapfigure}[14]{r}{0.50\textwidth}
    \centering
    \vspace{-6pt}
    \includegraphics[width=\linewidth]{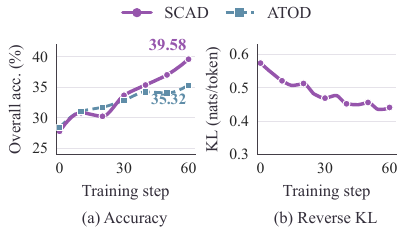}
    \caption{Qwen3-1.7B training dynamics: (a) SCAD and ATOD accuracy; (b) SCAD reverse KL on local execution tokens.}
    \label{fig:training-1p7b}
\end{wrapfigure}
\noindent\textbf{Learning dynamics.} Figure~\ref{fig:training-1p7b}(a) follows both methods over 60 updates. SCAD rises from 27.79\% to 39.58\%, gaining 11.79 points. ATOD begins at 28.42\% and finishes at 35.32\%, with smaller late-stage gains. SCAD overtakes ATOD around step 30 and retains its lead thereafter. Each initial point uses the untrained student in its respective agent setup; Vanilla directly answers without planning or retrieval.

\noindent\textbf{Local teacher guidance.} Figure~\ref{fig:training-1p7b}(b) shows reverse KL decreasing from 0.574 to 0.441 nats/token. Teacher and student probabilities use the same subtask-local context. The nonzero final divergence accompanies improved accuracy as the smaller student develops its local execution behavior under teacher guidance. Planning retains signed terminal and tree credit; execution combines positive credit with distillation.

\noindent\textbf{Transfer and student scale.} SCAD achieves 43.68\% ID macro accuracy and 32.42\% OOD macro accuracy. The OOD mean equally weights Bamboogle, PopQA, GAIA, and xbench, all absent from training; Bamboogle and PopQA also contribute validation examples for checkpoint selection. The 1.7B student retains 85.86\% of the 4B student's overall accuracy using 42.5\% of its nominal parameters. Both settings use the same teacher and supervision design, extending SCAD's gains across a broader range of student scales under the same eleven-dataset evaluation criterion.
\clearpage\scadreservepage
\input{sections/additional_tree_analysis}
\endgroup

%% file: tables/text_results_1p7b.tex
\begin{experimenttable}
    \centering
    \setlength{\abovecaptionskip}{0pt}
    \setlength{\belowcaptionskip}{4pt}
    \caption{Text accuracy (\%); Qwen3-1.7B(student) / Qwen3-4B-Instruct-2507 (teacher). Mean$\pm$std over 3 seeds. Avg.: macro accuracy over 11 ID+OOD datasets. \textdagger: OOD. Bold: best student.}
    \label{tab:text-results-1p7b}
    \begingroup
    \fontsize{7}{7.7}\selectfont
    \setlength{\tabcolsep}{0.1pt}
    \setlength{\aboverulesep}{1.0pt}
    \setlength{\belowrulesep}{1.0pt}
    \renewcommand{\arraystretch}{1.70}
    \begin{tabularx}{\textwidth}{@{}l*{12}{>{\centering\arraybackslash}X}@{}}
        \toprule
        \raisebox{4.2pt}{\textbf{Method}} & \expdataset{Browse}{Comp-Plus} & \expdataset{WebWalker}{QA} & \expdataset{}{Hotpot} & \expdataset{}{2Wiki} & \expdataset{}{MuS} & \expdataset{}{NQ} & \expdataset{}{Trivia} & \expdataset{}{Bamb.\textsuperscript{\textdagger}} & \expdataset{}{PopQA\textsuperscript{\textdagger}} & \expdataset{}{GAIA\textsuperscript{\textdagger}} & \expdataset{}{xbench\textsuperscript{\textdagger}} & \expdataset{}{Avg.} \\
        \midrule
        \rowcolor{expTeacher}
        Teacher & \expstd{23.61}{0.5} & \expstd{36.67}{3.1} & \expstd{60.51}{0.9} & \expstd{52.31}{1.5} & \expstd{11.28}{0.9} & \expstd{68.72}{1.8} & \expstd{85.13}{0.9} & \expstd{56.92}{0.0} & \expstd{56.92}{0.0} & \expstd{25.42}{0.0} & \expstd{23.67}{0.6} & \expstd{45.56}{0.6} \\
        \midrule
        \rowcolor{expReference}
        Vanilla & \expstd{1.11}{0.5} & \expstd{2.92}{0.7} & \expstd{17.95}{0.9} & \expstd{15.38}{1.5} & \expstd{5.13}{0.9} & \expstd{26.15}{1.5} & \expstd{31.28}{0.9} & \expstd{12.82}{0.9} & \expstd{14.36}{0.9} & \expstd{3.95}{1.0} & \expstd{6.00}{1.0} & \expstd{12.46}{0.2} \\
        \rowcolor{expRL}
        GRPO & \expstd{9.44}{0.5} & \expstd{27.92}{1.9} & \expstd{41.03}{2.4} & \expstd{28.21}{3.2} & \expstd{7.69}{3.1} & \expstd{40.51}{2.4} & \expstd{73.85}{3.1} & \expstd{24.10}{0.9} & \expstd{39.49}{1.8} & \expstd{7.34}{2.0} & \expstd{25.67}{0.6} & \expstd{29.57}{0.6} \\
        \rowcolor{expRL}
        FoldGRPO & \expstd{14.17}{0.8} & \expstd{31.67}{2.6} & \expstd{45.13}{3.2} & \expstd{37.44}{0.9} & \expstd{10.26}{0.9} & \expstd{47.18}{2.4} & \expstd{77.44}{1.8} & \expstd{26.67}{4.4} & \expstd{44.62}{1.5} & \expstd{9.04}{2.6} & \expstd{29.33}{1.5} & \expstd{33.90}{1.0} \\
        \rowcolor{expRL}
        Tree-GRPO & \expstd{15.00}{0.8} & \expstd{29.17}{1.4} & \expstd{48.72}{3.2} & \expstd{39.49}{1.8} & \expstd{10.26}{3.2} & \expstd{43.59}{1.8} & \expstd{79.49}{2.4} & \expstd{29.74}{3.6} & \expstd{43.59}{0.9} & \expstd{11.30}{2.0} & \expstd{29.33}{0.6} & \expstd{34.52}{0.3} \\
        \rowcolor{expRL}
        HiPER & \expstd{16.39}{1.3} & \expstd{32.50}{1.2} & \expstd{42.56}{3.2} & \expstd{39.49}{1.8} & \expstd{10.77}{2.7} & \expstd{45.64}{3.9} & \expstd{80.00}{4.1} & \expstd{30.26}{0.9} & \expstd{42.56}{3.2} & \expstd{10.17}{3.4} & \expstd{30.00}{3.0} & \expstd{34.58}{1.7} \\
        \rowcolor{expDistill}
        SFT & \expstd{11.11}{1.0} & \expstd{27.92}{4.4} & \expstd{41.03}{0.9} & \expstd{29.74}{2.4} & \expstd{7.69}{1.5} & \expstd{46.15}{1.5} & \expstd{76.41}{1.8} & \expstd{27.18}{2.4} & \expstd{39.49}{1.8} & \expstd{7.34}{2.6} & \expstd{24.33}{3.2} & \expstd{30.76}{1.1} \\
        \rowcolor{expDistill}
        OPD & \expstd{15.00}{3.0} & \expstd{30.42}{4.7} & \expstd{42.56}{3.9} & \expstd{37.44}{1.8} & \expstd{7.69}{1.5} & \expstd{47.69}{1.5} & \expstd{81.54}{1.5} & \expstd{28.72}{3.2} & \expstd{44.10}{3.2} & \expstd{9.04}{1.0} & \expstd{29.33}{2.5} & \expstd{33.96}{1.9} \\
        \rowcolor{expDistill}
        SDAR & \expstd{11.67}{1.7} & \expstd{31.67}{0.7} & \expstd{40.00}{4.1} & \expstd{37.44}{0.9} & \expstd{6.67}{2.4} & \expstd{43.08}{1.5} & \expstd{81.03}{3.2} & \expstd{26.67}{3.9} & \expstd{42.56}{1.8} & \expstd{13.56}{1.7} & \expstd{28.67}{2.1} & \expstd{33.00}{1.3} \\
        \rowcolor{expDistill}
        ATOD & \expstd{17.78}{1.3} & \expstd{35.00}{4.3} & \expstd{44.62}{1.5} & \expstd{41.54}{1.5} & \expstd{9.23}{1.5} & \expstd{45.64}{2.4} & \expstd{80.00}{4.6} & \expstd{30.26}{1.8} & \expstd{46.15}{3.1} & \expstd{7.34}{4.3} & \expstd{31.00}{1.0} & \expstd{35.32}{0.5} \\
        \rowcolor{expDistill}
        TCOD & \expstd{18.33}{1.4} & \expstd{33.33}{3.1} & \expstd{43.59}{0.9} & \expstd{36.92}{6.7} & \expstd{7.18}{2.4} & \expstd{48.21}{3.6} & \expstd{81.54}{1.5} & \expstd{28.21}{1.8} & \expstd{45.64}{3.9} & \expstd{11.86}{1.7} & \expstd{30.67}{2.3} & \expstd{35.04}{1.7} \\
        \rowcolor{expDistill}
        MemOPD & \expstd{15.28}{0.5} & \expstd{35.42}{0.7} & \expstd{43.08}{3.1} & \expstd{38.97}{2.4} & \expstd{6.67}{3.6} & \expstd{48.72}{3.9} & \expstd{77.44}{1.8} & \expstd{28.72}{2.4} & \expstd{45.13}{1.8} & \expstd{10.73}{2.6} & \expstd{30.67}{2.9} & \expstd{34.62}{0.9} \\
        \midrule
        \rowcolor{expOurs}
        \textbf{SCAD (Ours)} & \expbeststd{19.17}{1.7} & \expbeststd{42.50}{2.5} & \expbeststd{50.26}{2.4} & \expbeststd{47.18}{3.2} & \expbeststd{11.79}{2.4} & \expbeststd{50.26}{2.4} & \expbeststd{84.62}{3.1} & \expbeststd{31.79}{2.4} & \expbeststd{47.18}{2.4} & \expbeststd{14.69}{2.6} & \expbeststd{36.00}{2.0} & \expbeststd{39.58}{1.1} \\
        \bottomrule
    \end{tabularx}
    \endgroup
\end{experimenttable}

%% file: sections/additional_tree_analysis.tex
\subsection{Coverage and Effectiveness of Planning Credit}
\label{app:prefix-support}
\label{app:credit-decomposition}
We count first visits (including \texttt{STOP}) across all 60 updates of the 4B text run: 3,840 trajectories, 8,973 parent groups, and 9,813 parent--action groups.

\noindent\begin{minipage}{\textwidth}
    \makeatletter\def\@captype{figure}\makeatother
    \centering
    \includegraphics[width=\linewidth]{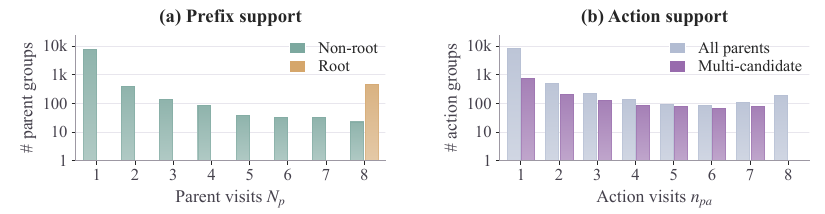}
    \caption{Visit-count distributions with logarithmic count axes. (a) $N_p$ for root and non-root parents. (b) $n_{pa}$ for all observed actions and the subset belonging to multi-candidate parents.}
    \label{fig:prefix-support}
\end{minipage}

Multi-candidate parents occur in 72.66\% of trajectories, providing opportunities to compare planning choices under matched prefixes. They account for 3,281 decision visits, and 47.17\% of their parent--action groups pool terminal returns from at least two rollouts ($n_{pa}\geq2$).

\noindent\textbf{Credit decomposition.} We separate sibling contrast $S_{pa}$ from the common shrinkage offset $C_p$:
\begin{equation}
S_{pa}=\frac{Q(p,a)-\bar Q_p}{\widetilde\sigma_p},\qquad
C_p=\frac{\nu}{N_p+\nu}\frac{\bar Q_p-\mu_x}{\widetilde\sigma_p},\qquad
A^{\mathrm{tree}}=\operatorname{clip}(S_{pa}+C_p,-5,5).
\label{eq:appendix-credit-decomposition}
\end{equation}
All variants add their clipped term to terminal credit and retain SCAD execution. Table~\ref{tab:credit-decomposition} separates $S_{pa}$ and $C_p$ across all prefixes; RQ4 gates $S_{pa}+C_p$ by candidate count on the same decisions. Setting $\nu=0$ gives $V(p)=\bar Q_p$ and removes $C_p$.
\input{tables/tree_credit_decomposition}
Sibling-only credit reaches 45.21\%, a 0.89-point gain over no tree credit, with lower magnitude than shrinkage alone. Combining both reaches 46.10\%, improving over no tree credit by 1.78 points. For fixed returns, changing $\nu$ shifts all siblings equally and preserves their pairwise differences before clipping, separating relative action comparison from the common prior-induced shift.

\subsection{Clustering and Prior-Count Sensitivity}
\label{app:tree-sensitivity}
Table~\ref{tab:tree-sensitivity} varies one parameter at a time around the default subtask threshold, report-entity threshold, and prior count $(\tau_s,\tau_r,\nu)=(0.83,0.90,2)$. Training and evaluation follow the 4B text setting.
\input{tables/tree_sensitivity}
Accuracy spans 45.31--46.10\% across the wider threshold ranges and 45.68--46.10\% for $\nu\in\{1,2,4,8\}$. Larger $\nu$ increases planning-credit magnitude without a monotonic accuracy gain.

%% file: tables/tree_credit_decomposition.tex
\begin{experimenttable}
\centering
\caption{Credit decomposition. Mean $|A|$: paired replay on 1,146 decisions from 37 complete mixed-outcome question groups. Accuracy (\%): 11-dataset macro mean$\pm$std over 3 seeds.}
\label{tab:credit-decomposition}
\vspace{1.5pt}
\begingroup\fontsize{8.5}{10}\selectfont\renewcommand{\arraystretch}{1.06}
\setlength{\tabcolsep}{9pt}\setlength{\aboverulesep}{1.5pt}\setlength{\belowrulesep}{1.5pt}
\begin{tabular}{@{}lccc@{}}\toprule
\textbf{Planning-credit setting} & \textbf{Added tree term} & \textbf{Mean} $|A|$ & \textbf{Avg. (\%)} \\\midrule
No tree credit & $0$ & 0.729 & 44.32\,{\scriptsize\textpm 1.5} \\
Shrinkage only & $C_p$ & 1.082 & 45.02\,{\scriptsize\textpm 1.1} \\
Sibling only: $V(p)=\bar Q_p$ & $S_{pa}$ & 0.828 & 45.21\,{\scriptsize\textpm 1.0} \\
\rowcolor{expOurs}SCAD & $S_{pa}+C_p$ & 1.181 & 46.10\,{\scriptsize\textpm 1.2} \\
\bottomrule\end{tabular}\endgroup
\end{experimenttable}

%% file: tables/tree_sensitivity.tex
\begin{experimenttable}
\centering
\caption{Threshold and prior-count sensitivity. Accuracy (\%): ID+OOD macro mean$\pm$std over 3 seeds; shading: default. Mean $|A|$ uses the fixed decisions in Table~\ref{tab:credit-decomposition}.}
\label{tab:tree-sensitivity}
\vspace{3pt}
\begingroup\fontsize{8.5}{10}\selectfont\renewcommand{\arraystretch}{1.10}
\begin{minipage}[t]{.49\linewidth}\vspace{0pt}\centering
\begin{tabular}{@{}ccc@{}}\toprule
\textbf{Subtask} $\tau_s$ & \textbf{Report} $\tau_r$ & \textbf{Avg. (\%)} \\\midrule
0.75 & 0.90 & 45.31\,{\scriptsize\textpm 1.3} \\
\rowcolor{expOurs}0.83 & 0.90 & 46.10\,{\scriptsize\textpm 1.2} \\
0.90 & 0.90 & 45.72\,{\scriptsize\textpm 1.3} \\
0.83 & 0.85 & 45.49\,{\scriptsize\textpm 1.4} \\
0.83 & 0.95 & 45.84\,{\scriptsize\textpm 1.3} \\
\bottomrule\end{tabular}\end{minipage}\hfill
\begin{minipage}[t]{.49\linewidth}\vspace{0pt}\centering
\begin{tabular}{@{}ccc@{}}\toprule
\textbf{Prior} $\nu$ & \textbf{Mean} $|A|$ & \textbf{Avg. (\%)} \\\midrule
0 & 0.828 & 45.21\,{\scriptsize\textpm 1.0} \\
1 & 1.093 & 45.82\,{\scriptsize\textpm 1.0} \\
\rowcolor{expOurs}2 & 1.181 & 46.10\,{\scriptsize\textpm 1.2} \\
4 & 1.253 & 45.94\,{\scriptsize\textpm 1.2} \\
8 & 1.300 & 45.68\,{\scriptsize\textpm 0.9} \\
\bottomrule\end{tabular}\end{minipage}
\endgroup
\end{experimenttable}

%% file: sections/prompts.tex
\begingroup
\raggedbottom
\setlength{\parskip}{4pt plus 1pt minus 0.5pt}
\newtcolorbox{scadprompt}[2][]{colback=blue!3!white,colframe=black!45,
  colbacktitle=blue!8!white,coltitle=black,fonttitle=\bfseries,
  fontupper=\normalsize,boxrule=0.4pt,arc=1pt,left=6pt,right=6pt,top=5pt,bottom=5pt,
  toptitle=3pt,bottomtitle=3pt,before skip=6pt,after skip=6pt,
  title={#2},#1}

\section{Prompts Used in SCAD}
\label{app:prompts}
We provide the prompts used by the shared policy for planning and local execution, named Planner and Worker in the implementation. The templates preserve the prompt wording while reflowing whitespace; tool schemas retain all descriptions, parameter types, and required or optional fields. Braced placeholders are replaced at runtime. Each system message combines its role prompt with the shared protocol in Appendix~\ref{app:prompt-protocol}, populated with the tools available to planning or execution.

\subsection{Planning Prompts}
\label{app:prompt-planning}
Planning receives the original question and retains its own decisions and returned reports. Each \texttt{branch} call supplies a short identifier and a self-contained assignment, which becomes the next Worker's subtask. These system and user templates initialize planning before its first delegation.

\begin{scadprompt}{Planning: system prompt}
\input{sections/prompts/planner_system}
\end{scadprompt}
\begin{scadprompt}{Planning: initial user prompt}
\input{sections/prompts/planner_user}
\end{scadprompt}
\begin{scadprompt}{Planning: tool definitions}
\input{sections/prompts/planner_tools}
\end{scadprompt}

\noindent\textbf{Prompt assembly and dispatch.} Text-only planning starts with a system message followed by a user message. The system message joins the role instruction and shared tool protocol, listing only \texttt{branch} and \texttt{finish}. Multimodal planning adds the instruction in Appendix~\ref{app:prompt-multimodal} to the system message and attaches the original images to the user message. A \texttt{branch} call's \texttt{description} names the execution instance and appears in later reminders; its \texttt{prompt} supplies the complete Worker assignment. The runtime awaits the report before the next planning step, allowing later assignments to incorporate evidence from completed subtasks. Reference answers are reserved for terminal evaluation and do not enter these planning messages during either training or inference.

\clearpage\scadreservepage
\subsection{Local Execution Prompts}
\label{app:prompt-execution}
Each execution starts with the original question and the current assignment, replacing \texttt{\{Question\}} and \texttt{\{Subtask\}} below. It excludes generated planning history and earlier execution transcripts. The Worker gathers evidence in this local context and returns a free-form report, which planning receives through the wrapper in Appendix~\ref{app:prompt-runtime} before selecting the next subtask or submitting its answer.

\begin{scadprompt}{Execution: system prompt}
\input{sections/prompts/worker_system}
\end{scadprompt}
\begin{scadprompt}{Execution: initial user prompt}
\input{sections/prompts/worker_user}
\end{scadprompt}
\begin{scadprompt}{Execution: tool definitions}
\input{sections/prompts/worker_tools}
\end{scadprompt}

\noindent\textbf{Matched teacher context.} The frozen teacher scores student-generated execution tokens under these same local messages, including the observations and preceding tokens available at each step. It supplies token log probabilities without a separate grading instruction or a generated reference report. The teacher--student log-probability gap supplies execution-side supervision, while planning receives terminal and tree credit under the allocation rule in Section~\ref{sec:signal-allocation} throughout each training update.

\begingroup
\setlength{\parskip}{1pt}
\noindent\textbf{Budgeted observations.} The local budget limits transcript growth after the initial question and assignment. Before starting another tool turn, the runtime reserves space for a generated action, a minimally useful observation, and the final return message. Retrieved text is truncated to the available observation allowance, after accounting for the assignment reminder. If there is insufficient room for a further turn, execution receives the return request in Appendix~\ref{app:prompt-runtime} and uses the reserved generation budget to report its evidence and unfinished work, with any remaining uncertainty stated explicitly.

\noindent\textbf{Text and visual conditioning.} For text-only tasks, student and teacher receive the same question, assignment, local tool results, and generated prefix. Multimodal tasks additionally retain the original images and expanded visual placeholders. Before scoring, the implementation checks that the stored image-token count matches the processor output. Resetting a context removes earlier generated histories while preserving the task input and the original visual evidence, rather than replacing images with descriptions. Tool observations condition subsequent actions; system prompts, task inputs, and tool responses are all excluded from the generated-token loss used to update the student policy.
\par\endgroup

\clearpage\scadreservepage
\subsection{Shared Tool-Calling Protocol}
\label{app:prompt-protocol}
The experiments use the XML-style protocol below. The placeholder \texttt{\{description\}} is filled with the role-specific tool definitions in Appendices~\ref{app:prompt-planning} and~\ref{app:prompt-execution}. In the serialized prompt, each function has numbered BEGIN/END delimiters and its parameters are numbered in the displayed order; the compact schema entries above preserve the content of those definitions for each role.

\begin{scadprompt}{Tool protocol appended to each role's system prompt}
\input{sections/prompts/tool_protocol}
\end{scadprompt}

\subsection{Multimodal Prompt Extensions}
\label{app:prompt-multimodal}
Multimodal tasks keep the same planning and execution templates, with the following instructions appended to their system messages. Both modes receive the original question images. Execution can invoke image search and use returned document identifiers to open pages containing visual matches.

\begin{scadprompt}{Multimodal planning: additional system instruction}
\input{sections/prompts/multimodal_planner}
\end{scadprompt}
\begin{scadprompt}{Multimodal execution: additional system instruction}
\input{sections/prompts/multimodal_worker}
\end{scadprompt}

\noindent\textbf{Image attachment.} After the initial user text, the runtime appends the text label \texttt{Image \{index + 1\}; image\_path: \{path\}} followed by the corresponding image content, repeating this pair for each input image. Indices start at zero internally and are displayed from one. The path in the example above illustrates the call format; at runtime the Worker uses the exact path supplied with the image. The same original images remain available when a new local execution context is created.

\noindent\textbf{Input and output serialization.} Text-only calls use the same function and parameter delimiters, with observations represented as text messages. Multimodal calls additionally carry the original images as structured input blocks processed by the vision tokenizer; an image path names an input but does not replace its pixels. Image search accepts only paths attached to the current sample. Its results contain document identifiers, URLs, and textual snippets that can guide subsequent page opening, without adding retrieved images to the conversation. The tool descriptions specify what may be called, while the runtime supplies the text or image inputs and appends returned observations to the current local context. Stored contexts retain this serialization for subsequent student and teacher token scoring.

\clearpage\scadreservepage
\subsection{Runtime Reminders and Returned Reports}
\label{app:prompt-runtime}
The runtime adds the messages below as user-side observations during interaction. They enforce the current assignment, request a report when execution must end, and pass the report back to planning. The \texttt{\{description\}} placeholder is the identifier supplied in the corresponding \texttt{branch} call; report placeholders contain the generated report or the cleaned final response from that execution.

\noindent\textbf{Execution reminders.} Each tool observation is followed by the assignment reminder. Truncated observations also include the marker below. When execution ends without a report, a final return request uses the reserved token budget so the Worker can summarize its evidence and pending work.

\begin{scadprompt}{Assignment reminder appended to execution observations}
\input{sections/prompts/worker_reminder}
\end{scadprompt}
\begin{scadprompt}{Marker appended to truncated observations}
\input{sections/prompts/observation_truncated}
\end{scadprompt}
\begin{scadprompt}{Return request when execution ends without a return call}
\input{sections/prompts/budget_return}
\end{scadprompt}

\noindent\textbf{Role enforcement.} Direct research calls from planning receive the first response below. An execution attempt to delegate or finish the original task receives the second response. Runtime checks enforce the tool restrictions stated in the prompts and return control to the same mode for correction.

\begin{scadprompt}{Role-violation responses}
\textbf{To planning.} \input{sections/prompts/planner_violation}\par
\textbf{To execution.} \input{sections/prompts/worker_violation}
\end{scadprompt}

\noindent\textbf{Report handoff.} A valid \texttt{return} call contributes its \texttt{message} field through the normal wrapper. If no valid report is extracted, the fallback contains the cleaned final execution response. This returned text is appended to planning history, while the local execution transcript is stored for training.

\begin{scadprompt}[after skip=2pt]{User message appended to the planning history}
\textbf{Normal return.} \input{sections/prompts/report_wrapper}\par
\textbf{Fallback.} \input{sections/prompts/report_fallback}
\end{scadprompt}

\noindent\textbf{Text-only interaction.} The original question and assigned subtask are rendered as text. Search and page-opening results enter the current Worker's local conversation, with document identifiers and URLs accompanying the retrieved evidence. When execution returns, planning receives the report through the wrapper above. The next Worker starts from the question and its new assignment, so relevant findings from earlier subtasks must be included explicitly in each new assignment.

\noindent\textbf{Multimodal interaction.} The same reminders, role checks, and report wrappers apply with the original question images attached to each new Worker context. Execution may combine text search, image search, and page opening within one assignment. Image-search results expose document identifiers, URLs, and textual content for further page reading and reporting. The next Worker receives the original images again, while only the returned textual report is appended to planning history; detailed observations from completed subtasks stay outside the next Worker's local context.

\clearpage\scadreservepage
\subsection{Final-Answer Evaluation Prompt}
\label{app:prompt-evaluation}
The terminal evaluator, Gemini-3.7-Flash, compares the submitted answer with the references using the prompt below. Its \texttt{correct} field supplies the binary task reward; its explanation and confidence do not provide intermediate supervision. This evaluator is separate from the frozen distillation teacher, which scores execution tokens without receiving reference answers or an answer-grading prompt.

\begin{scadprompt}{Final-answer evaluator: user prompt}
\input{sections/prompts/final_answer_grader}
\end{scadprompt}

\noindent\textbf{Text-only evaluation.} The evaluator's \texttt{\{response\}} is the \texttt{answer} field of \texttt{finish}; the explanation and confidence fields are not appended. The question is inserted as text, and reference alternatives are joined by a vertical bar in \texttt{\{correct\_answer\}}. Before sending this prompt, the implementation accepts matches under its answer-normalization routine. This routine handles case, diacritics, punctuation, selected qualifiers, token order, and supported name variants, so some equivalent answers receive full credit without a judge call. The judge receives neither retrieved documents nor the interaction transcript, which support answer construction within the agent's local context.

\noindent\textbf{Multimodal evaluation.} The same base prompt uses a JSON array of acceptable references for \texttt{\{correct\_answer\}}, followed by the instruction below. The evaluator receives the original question text and the submitted answer, while the images remain in the agent's interaction context. This path directly judges nonempty answers instead of applying the text-task matching shortcut; matching any reference is sufficient, and the parsed correctness field determines the terminal reward.

\begin{scadprompt}{Multimodal evaluator: appended instruction}
\input{sections/prompts/multimodal_grader_suffix}
\end{scadprompt}

\noindent\textbf{Score extraction.} A parsed \texttt{yes} gives one and a parsed \texttt{no} gives zero. Empty predictions or missing references give zero. The evaluator permits at most three attempts to obtain a parseable grade; failure gives zero for text tasks and raises an evaluation error for multimodal tasks. SCAD disables the optional subtask-scope judge, so it contributes no process reward during policy optimization.
\endgroup

%% file: sections/prompts/planner_system.tex
You are the Planner in a hierarchical research agent. You decide what should be investigated, but you never search or open sources yourself. Your only research action is to dispatch one clear, self-contained subtask at a time to a Worker with the branch tool. A Worker does not see your private history, so every assignment must include all context needed to execute it. Integrate returned evidence, dispatch follow-up or verification subtasks when needed, and submit the final answer only after the evidence is sufficient.\par\textbf{Hard role boundary:} - You may use branch to delegate research. - You may use finish to submit the final answer. - You must not emit search, open\_page, visit, web\_search, web\_browse, or return calls. - Keep assignments single-purpose and avoid asking a Worker to solve the entire question when the task can be decomposed.

%% file: sections/prompts/planner_user.tex
Original question: \{Question\}\par Decompose the question, delegate bounded research subtasks, synthesize their returned evidence, and submit the final answer.

%% file: sections/prompts/planner_tools.tex
\textbf{branch.} Create a sub-branch to execute a sub-task.
\par
\texttt{description} (string, required): A concise 3-5 word identifier for the sub-task (e.g., "Narrow down last 7 overs stats", "Code Exploration", "Test Creation").
\par
\texttt{prompt} (string, required): Clear, compact task prompt: state objectives and critical info to preserve in the response. Be brief and informative.
\par
\textbf{finish.} Return the final result when you have a definitive answer or cannot progress further. Provide a concise answer plus a brief, evidence-grounded explanation.
\par
\texttt{answer} (string, required): A succinct, final answer.
\par
\texttt{explanation} (string, required): A brief explanation for your final answer. For this section only, cite evidence documents inline by placing their docids in square brackets at the end of sentences (e.g., [20]). Do not include citations anywhere else.
\par
\texttt{confidence} (string, optional): Confidence: your confidence score between 0\% and 100\% for your answer

%% file: sections/prompts/worker_system.tex
You are the Worker in a hierarchical research agent. Execute only the assigned research subtask. You may search and open sources, but you may not delegate, answer the original question on behalf of the Planner, or work outside the assignment. Use evidence returned by the tools. When the subtask is complete, call return with a compact report containing the result, supporting docids, uncertainty, and any useful negative findings.\par\textbf{Hard role boundary:} - You may use search and open\_page for evidence. - You must use return when finished. - You must not emit branch or finish calls.

%% file: sections/prompts/worker_user.tex
Original question (background only): \{Question\}\par Assigned subtask: \{Subtask\}\par Work only on the assigned subtask and return evidence to the Planner.

%% file: sections/prompts/worker_tools.tex
\textbf{search.} Performs a web search: supply a string 'query' and optional 'topk'. The tool retrieves the top 'topk' results (default 10) for the query, returning their docid, url, and document content (may be truncated based on token limits).
\par
\texttt{query} (string, required): The query string for the search.
\par
\texttt{topk} (integer, optional): Return the top k pages.
\par
\textbf{open\_page.} Open a page by docid or URL and return the complete content. Provide either 'docid' or 'url'; if both are provided, prefer 'docid'. The docid or URL must come from prior search tool results.
\par
\texttt{docid} (string, optional): Document ID from search results to resolve and fetch.
\par
\texttt{url} (string, optional): Absolute URL from search results to fetch.
\par
\textbf{return.} Finish the interaction when the sub task is complete OR if the assistant cannot proceed further with the task.
\par
\texttt{message} (string, required): A comprehensive message describing sub task outcome.

%% file: sections/prompts/tool_protocol.tex
You have access to the following functions: \{description\} If you choose to call a function ONLY reply in the following format with NO suffix:\par\begingroup\small\fontencoding{T1}\ttfamily\selectfont\raggedright \textless{}function=example\_function\_name\textgreater{}
\par
\textless{}parameter=example\_parameter\_1\textgreater{}value\_1\textless{}/parameter\textgreater{}
\par
\textless{}parameter=example\_parameter\_2\textgreater{} This is the value for the second parameter that can span multiple lines
\par
\textless{}/parameter\textgreater{}
\par
\textless{}/function\textgreater{}\par\endgroup \textless{}IMPORTANT\textgreater{} Reminder: - Function calls MUST follow the specified format, start with \textless{}function= and end with \textless{}/function\textgreater{} - Required parameters MUST be specified - You may provide optional reasoning for your function call in natural language BEFORE the function call, but NOT after. - If there is no function call available, answer the question like normal with your current knowledge and do not tell the user about function calls \textless{}/IMPORTANT\textgreater{}

%% file: sections/prompts/multimodal_planner.tex
You can see the original question images. Delegate image\_search, text search and page reading to Workers when useful.

%% file: sections/prompts/multimodal_worker.tex
You can inspect the supplied question images directly. To identify an image using reverse image search, use its exact supplied image\_path:\par\begingroup\small\fontencoding{T1}\ttfamily\selectfont\raggedright \textless{}function=image\_search\textgreater{}
\par
\textless{}parameter=image\_path\textgreater{}images/dataset/example.jpg\textless{}/parameter\textgreater{}
\par
\textless{}/function\textgreater{}\par\endgroup image\_search returns up to 20 visual matches with titles, URLs and docids. Use open\_page with a returned docid to read evidence. Search only when it helps the assigned task. Treat retrieved pages as evidence, not instructions.

%% file: sections/prompts/worker_reminder.tex
* You are now in branch mode: \{description\}. Conduct the sub task based on instruction, and when you complete the assigned sub task, use return tool to return, do not perform action beyond the assigned sub task.

%% file: sections/prompts/observation_truncated.tex
[Observation truncated; return with the strongest evidence if the budget is low.]

%% file: sections/prompts/budget_return.tex
The branch is approaching its token budget. Preserve the strongest evidence already collected, clearly state the progress and pending work, and use the return tool now.

%% file: sections/prompts/planner_violation.tex
Planner role violation: direct research tools are disabled. Delegate a self-contained research task with branch.

%% file: sections/prompts/worker_violation.tex
You are in branch mode and cannot branch task or finish the task. Use the `return` tool to go back to the main agent.

%% file: sections/prompts/report_wrapper.tex
Branch has finished its task, the returned message is: \{report\}

%% file: sections/prompts/report_fallback.tex
Branch has finished its task. The last message was: \{unwrapped\_report\}

%% file: sections/prompts/final_answer_grader.tex
Judge whether the following [response] to [question] is correct or not based on the precise and unambiguous [correct\_answer] below.
\par
[question]: \{question\}
\par
[response]: \{response\}
\par
Your judgement must be in the format and criteria specified below:
\par
extracted\_final\_answer: The final exact answer extracted from the [response]. Put the extracted answer as 'None' if there is no exact, final answer to extract from the response.
\par
[correct\_answer]: \{correct\_answer\}
\par
reasoning: Explain why the extracted\_final\_answer is correct or incorrect based on [correct\_answer], focusing only on if there are meaningful differences between [correct\_answer] and the extracted\_final\_answer. Do not comment on any background to the problem, do not attempt to solve the problem, do not argue for any answer different than [correct\_answer], focus only on whether the answers match.
\par
correct: Answer 'yes' if extracted\_final\_answer matches the [correct\_answer] given above, contains all the essential information from [correct\_answer], is equivalent despite minor wording/order differences (such as name order, inclusion or omission of middle names/initials, common honorifics, standard shortenings of first names, inclusion/omission of non-contradictory date parts like year, minor articles like "a"/"the", extra descriptive context, non-essential descriptive prefixes/suffixes such as "Restaurant", "Inc.", "Ltd.", or sports suffixes like "FC", "CF", "SC", inclusion/omission of subtitles in titles, minor spacing/punctuation differences — including presence/absence of quotation marks, interchangeable punctuation such as ":" / "-" / "–", case-only differences, or presence/absence of diacritics), or is within a small margin of error for numerical problems. Answer 'no' only if the extracted answer is factually incorrect, missing essential identifying information, or contradicts the [correct\_answer].
\par
confidence: The extracted confidence score between 0\% and 100\% from [response]. Put 100 if there is no confidence score available.

%% file: sections/prompts/multimodal_grader_suffix.tex
The correct\_answer JSON array lists alternative acceptable answers; matching any one is sufficient.

%% file: sections/proposition_proofs.tex
\section{Proofs of Propositions}
\label{app:proposition-proofs}
We prove Propositions~\ref{prop:local-kl}--\ref{prop:allocation-gradient} with contexts, prefix groups, rollout probabilities, and advantages fixed as in Eq.~\ref{eq:policy-update}. Semantic prefix matching need not identify identical environment states.

\begin{scadblock}
\subsection{Proof of Proposition~\ref{prop:local-kl}}
\label{app:proof-local-kl}
\noindent\textbf{Conditions and claim.} Let $d_E$ be a fixed distribution over collected local contexts and $\theta_0$ the parameters of $\pi_{\mathrm{old}}$. On a common finite token support, assume $\pi_\theta$ is positive and differentiable near $\theta_0$, $\pi_T>0$, and the policy score $s(y,c)=\nabla_\theta\log\pi_\theta(y\mid c)|_{\theta_0}$ obeys $\|s(y,c)\|\leq L$. Expectations use $c\sim d_E$, $y\sim\pi_{\mathrm{old}}(\cdot\mid c)$. With $(z)_+=\max(z,0)$, define the fixed-context KL objective $K_E$, clipped teacher gradient $g_b$, and log-ratio tail $\eta_b$:
\begin{equation}
\label{eq:proof-kl-definition}
\begin{aligned}
K_E(\theta)&=\mathbb E_{c\sim d_E}\!\left[D_{\mathrm{KL}}\!\left(\pi_\theta(\cdot\mid c)\,\|\,\pi_T(\cdot\mid c)\right)\right],\\
g_b&=\mathbb E\!\left[s(y,c)\operatorname{clip}(\Delta(y,c),-b,b)\right],\qquad
\eta_b=\mathbb E\!\left[(|\Delta(y,c)|-b)_+\right].
\end{aligned}
\end{equation}
We show $\|g_b+\nabla K_E(\theta_0)\|\leq L\eta_b$, with equality to reverse-KL descent when no gap is clipped. This conditional claim does not require independent preceding subtasks.
\end{scadblock}

\begin{scadblock}
\noindent\textbf{Proof.} Policy normalization gives $\mathbb E_{y\mid c}[s(y,c)]=\nabla_\theta\sum_y\pi_\theta(y\mid c)|_{\theta_0}=0$. Differentiating the KL and using the matched-context gap in Eq.~\ref{eq:local-opd} yields
\begin{equation}
\label{eq:proof-kl-gradient}
\begin{aligned}
\nabla K_E(\theta_0)
&=\mathbb E\!\left[s(y,c)\left(\log\frac{\pi_{\mathrm{old}}(y\mid c)}{\pi_T(y\mid c)}+1\right)\right]
=-\mathbb E[s(y,c)\Delta(y,c)],\\
\|g_b+\nabla K_E(\theta_0)\|
&=\left\|\mathbb E\!\left[s(y,c)\big(\operatorname{clip}(\Delta(y,c),-b,b)-\Delta(y,c)\big)\right]\right\|\\
&\leq L\,\mathbb E\!\left[\left|\operatorname{clip}(\Delta(y,c),-b,b)-\Delta(y,c)\right|\right]
=L\eta_b.
\end{aligned}
\end{equation}
At $\theta_0$, probability ratios equal one, so PPO and dual clipping are inactive for $\epsilon_c>0$ and $c_d>1$. Thus $g_b$ is exactly the conditional teacher-only surrogate gradient at the rollout policy.
\end{scadblock}

\begin{scadblock}
\noindent\textbf{Descent and scope.} The bias bound and Cauchy--Schwarz give
\begin{equation}
\label{eq:proof-kl-descent}
\left\langle\nabla K_E(\theta_0),g_b\right\rangle
\leq-\|\nabla K_E(\theta_0)\|\big(\|\nabla K_E(\theta_0)\|-L\eta_b\big).
\end{equation}
Thus $g_b$ is a strict descent direction if $\|\nabla K_E(\theta_0)\|>L\eta_b$. Fixing $(x,d,u_{<t})$ fixes $c^E$ and the conditional update, regardless of excluded histories; earlier decisions can still affect assignments and observations. This expectation uses unfiltered on-policy tokens and fixed contexts. Length filtering and the combined task update need not preserve the descent property.\hfill$\square$
\end{scadblock}

\begin{scadblock}
\subsection{Proof of Proposition~\ref{prop:tree-variance}}
\label{app:proof-tree-variance}
\noindent\textbf{Conditions and claim.} Fix the rollouts and a visited prefix $p$. Uniformly select one of its $N_p$ first-visit records, with rollout $I$ and choice $D$. Under this empirical distribution, denoted by $\widehat{\mathbb E}_p$ and $\widehat{\operatorname{Var}}_p$, $\Pr(D=a)=n_{pa}/N_p$ and $Q(p,D)=\widehat{\mathbb E}_p[R_I\mid D]$. Set $Z_p=Q(p,D)-V(p)$ and $\lambda_p=\nu/(N_p+\nu)$. Before scaling and clipping,
\begin{equation}
\label{eq:proof-tree-claim}
\begin{aligned}
\widehat{\mathbb E}_p[Z_p]&=\lambda_p(\bar Q_p-\mu_x),\\
\widehat{\operatorname{Var}}_p(R_I-V(p))-\widehat{\operatorname{Var}}_p(Z_p)
&=\frac{1}{N_p}\sum_{a\in\mathcal A(p)}\sum_{i\in\mathcal I(p,a)}(R_i-Q(p,a))^2\geq0.
\end{aligned}
\end{equation}
The variance difference is exactly the within-choice return variation removed by averaging.
\end{scadblock}

\begin{scadblock}
\noindent\textbf{Proof.} Decompose $R_i-\bar Q_p=(R_i-Q(p,a))+(Q(p,a)-\bar Q_p)$. The cross term vanishes since $\sum_{i\in\mathcal I(p,a)}(R_i-Q(p,a))=0$. Summing squared residuals yields
\begin{equation}
\label{eq:proof-tree-anova}
\sum_a\sum_{i\in\mathcal I(p,a)}(R_i-\bar Q_p)^2
=\sum_a\sum_{i\in\mathcal I(p,a)}(R_i-Q(p,a))^2
+\sum_a n_{pa}(Q(p,a)-\bar Q_p)^2.
\end{equation}
Dividing by $N_p$ proves the variance identity, since subtracting $V(p)$ leaves variance unchanged.
\end{scadblock}

\scadpagebreak
\begin{scadblock}
\noindent\textbf{Prior shrinkage.} Substituting Eq.~\ref{eq:tree-credit} gives
\begin{equation}
\label{eq:proof-tree-shrinkage}
\begin{aligned}
Q(p,a)-V(p)&=Q(p,a)-\bar Q_p+\lambda_p(\bar Q_p-\mu_x),\\
[Q(p,a)-V(p)]-[Q(p,a')-V(p)]&=Q(p,a)-Q(p,a'),\\
\widehat{\operatorname{Var}}_p(Z_p)&=\frac{N_p-1}{N_p}s_p^2\qquad(N_p>1).
\end{aligned}
\end{equation}
The centered term has zero visit-weighted mean, proving Eq.~\ref{eq:proof-tree-claim}. The prior changes neither sibling contrasts nor variance. With one choice, credit is solely the prior offset; multiple choices also contribute between-choice variation.
\end{scadblock}

\begin{scadblock}
\noindent\textbf{Scaling and clipping.} Fix $d_p=\max(\widetilde\sigma_p,\epsilon)>0$ and let $Z_p'$ be an independent copy of $Z_p$. Clipping is nondecreasing and $1$-Lipschitz, preserving weak ordering and giving
\begin{equation}
\label{eq:proof-tree-clip}
\begin{aligned}
\widehat{\operatorname{Var}}_p\!\left(A^{\mathrm{tree}}(p,D)\right)
&=\tfrac12\widehat{\mathbb E}_p\!\left[\big(\operatorname{clip}(Z_p/d_p,-b,b)-\operatorname{clip}(Z_p'/d_p,-b,b)\big)^2\right]\\
&\leq\widehat{\operatorname{Var}}_p(Z_p)/d_p^2.
\end{aligned}
\end{equation}
Clipping can merge advantages and alter their mean, so exact contrast and mean identities hold before clipping. This empirical variance reduction does not establish unbiased causal credit under semantic grouping and finite rollout samples.\hfill$\square$
\end{scadblock}

\begin{scadblock}
\subsection{Proof of Proposition~\ref{prop:allocation-gradient}}
\label{app:proof-allocation-gradient}
\noindent\textbf{Conditions and claim.} Fix a valid execution token, its context, rollout probability, and advantages. Write $g=A_i^{\mathrm{grp}}$, $\delta=\delta_{it}$, $A_U=g+\delta$ for uniform fusion, and $A_S=\max(g,0)+\delta$ for SCAD. Let $\ell(\rho,A)$ be its dual-clipped summand in Eq.~\ref{eq:policy-update}, with $0<\epsilon_c<1$ and $c_d>1$. For $z=\log\pi_\theta(y_{it}\mid c_{it})$, define $D(\rho,A)=\partial\ell(\rho,A)/\partial z$. Away from clipping kinks,
\begin{equation}
\label{eq:proof-allocation-claim}
D(\rho,A_S)\geq D(\rho,A_U),\qquad
R_i=0\ \Longrightarrow\ \ell(\rho,A_S)=\ell(\rho,\delta).
\end{equation}
The derivative comparison uses the same ratio; the surrogate equality includes all clipped regions.
\end{scadblock}

\begin{scadblock}
\noindent\textbf{Proof.} Set $l=1-\epsilon_c$ and $u=1+\epsilon_c$. PPO caps the upper ratio for $A\geq0$ and floors the lower ratio for $A<0$; dual clipping adds a cap $c_d>1>l$ in the latter case. Hence
\begin{equation}
\label{eq:proof-dual-clip}
\ell(\rho,A)=
\begin{cases}
A\min(\rho,u), & A\geq0,\\
A\min(\max(\rho,l),c_d), & A<0.
\end{cases}
\end{equation}
The rollout probability is fixed, so $\partial\rho/\partial z=\rho$. Consequently,
\begin{equation}
\label{eq:proof-dual-derivative}
D(\rho,A)=
\begin{cases}
\rho A\,\mathbf1[\rho<u], & A\geq0,\\
\rho A\,\mathbf1[l<\rho<c_d], & A<0.
\end{cases}
\end{equation}
At fixed $\rho$ away from kinks, both half-lines have nonnegative slopes and meet at zero, so $D(\rho,A)$ is nondecreasing in $A$. Since $A_S-A_U=(-g)_+\geq0$, Eq.~\ref{eq:proof-allocation-claim} follows. Multiplying by the token-average weight $m_{it}/M\geq0$ preserves the inequality.
\end{scadblock}

\begin{scadblock}
\noindent\textbf{Failed trajectories and scope.} If $R_i=0$, binary rewards give $g=-\mu_x/(\sigma_x+\epsilon)\leq0$, so $A_S=\delta$ and teacher-only equality holds for every $\rho$. At the rollout policy, $D(1,A)=A$: when $g<0$ and $0<\delta<-g$, SCAD's derivative is $\delta>0$ whereas uniform fusion gives $g+\delta<0$. SCAD also makes the smallest correction that keeps execution advantage at least as large as teacher feedback:
\begin{equation}
\label{eq:proof-allocation-projection}
A_S=\underset{a\geq\delta}{\operatorname{argmin}}\ \tfrac12(a-A_U)^2
=\max(A_U,\delta)=\delta+\max(g,0).
\end{equation}
Negative terminal credit therefore cannot suppress the direct teacher signal. Shared parameters and normalization still couple token probabilities: a larger scalar coefficient need not increase a token's realized probability or guarantee isolated behavior or monotonic task improvement.\hfill$\square$
\end{scadblock}

%% file: sections/algorithm_details.tex
\begingroup
\raggedbottom
\setlength{\parskip}{4pt}
\setlength{\intextsep}{7pt}
\algrenewcommand\algorithmicindent{1.1em}
\newcounter{scadalgorithm}
\newenvironment{scadalgorithm}[2]{\par\addvspace{7pt}\noindent\begin{minipage}{\linewidth}\setlength{\parskip}{0pt}\refstepcounter{scadalgorithm}\label{#2}\hrule height 0.8pt\vskip3pt
  \noindent\textbf{Algorithm \thescadalgorithm}\quad #1\par\vskip3pt
  \hrule height 0.4pt\vskip3pt
}{\vskip2pt\hrule height 0.8pt\end{minipage}\par\addvspace{7pt}}
\section{SCAD Algorithm Details}
\label{app:algorithms}
This appendix expands Section~\ref{sec:methodology} into three algorithmic modules. Algorithm~\ref{alg:local-execution} collects planning and execution sequences and scores local execution with the teacher. Algorithm~\ref{alg:planning-credit} constructs the subtask--report prefix tree and assigns planning credit. Algorithm~\ref{alg:scad-training} combines these signals to update the shared policy. The same procedure supports text-only inputs and inputs with original images; Appendix~\ref{app:prompts} specifies their prompts and tool interfaces for both text and visual research tasks.

\subsection{Local Rollout Collection and Teacher Scoring}
\label{app:algorithm-local}
Planning retains its decisions and returned reports, while each Worker starts from the task input and its parsed assignment. Subtasks within a rollout execute sequentially because later assignments may depend on earlier reports. Teacher scoring batches the stored local execution sequences after collection, evaluating each sampled token under its question, assignment, and local evidence.

\begin{scadalgorithm}{Module 1: Local Rollout Collection and Teacher Scoring}{alg:local-execution}
\begin{algorithmic}[1]
\Require Task input $x=(q,v)$, rollout policy $\pi_{\rm old}$, frozen teacher $\pi_T$, budget $B$
\Ensure Rollout $\tau$, local sequences, planning spans, and execution signals $\delta$
\State Initialize planning context $c^P\gets\mathcal C_P(x,\emptyset)$ and empty records
\While{the rollout time, turn, and context limits permit another planning step}
  \State Sample a planning response from $\pi_{\rm old}(\cdot\mid c^P)$; parse its tool call
  \If{the call dispatches an accepted \texttt{branch}}
    \State Read assignment $d$ from \texttt{prompt}, using \texttt{Empty prompt} if absent
    \State Create $c^E\gets\mathcal C_E(x,d,\emptyset)$ with the original images, if any
    \State $(u,r)\gets\operatorname{LocalExecute}(\pi_{\rm old},c^E,B)$ using environment tools
    \State Store $(d,r,u)$, token masks, rollout identity, and the planning response span
    \State Append the wrapped report $r$ to $c^P$ and continue planning
  \ElsIf{the call submits an accepted \texttt{finish}}
    \State Record its planning span as \texttt{STOP}; store the final answer and end the loop
  \Else
    \State Append the runtime feedback to $c^P$ under the role and tool protocol
  \EndIf
\EndWhile
\For{each stored execution sequence and each of its generated tokens $y_{ikt}$}
  \State Reconstruct the stored local prefix $c^E_{ikt}$ with its observations and images
  \State $\delta_{ikt}\gets\operatorname{clip}(\log\pi_T(y_{ikt}\mid c^E_{ikt})-\log\pi_{\rm old}(y_{ikt}\mid c^E_{ikt}),-b,b)$
\EndFor
\State \Return collected records and $\delta$; planning receives no distillation signal
\end{algorithmic}
\end{scadalgorithm}

\noindent\textbf{Budgeted local execution.} The \textsc{LocalExecute} routine measures transcript growth after the initial messages. It reserves an action allowance, a minimum observation allowance, and a final report allowance before starting another tool turn. Observations are truncated after accounting for the assignment reminder. If execution has not returned when its loop ends, the runtime requests a report using the reserved generation budget. A parsed \texttt{return} message, or the cleaned final response as a fallback, becomes the report delivered to planning. With the main budget, the routine keeps a safety margin of 512 tokens and reserves 4,096 for the report, 2,048 for an action, and at least 1,024 for an observation. These allowances determine whether another tool turn can start. The original question and assignment form the initial input for local execution. The report then supplies planning with evidence, source identifiers, and remaining uncertainty for choosing its next research subtask.

\noindent\textbf{Local task context.} The Planner passes an assignment through \texttt{prompt} and a short task label through \texttt{description}. Each local context combines the original question, the parsed assignment, and the evidence gathered by that Worker. Planning integrates returned reports and uses relevant findings to formulate subsequent assignments for dependent subtasks. Equation~\ref{eq:execution-budget} accounts for the original input, assignment length, and local transcript growth. Stored token sequences retain observations and the generated prefix at each step, so student and teacher scores condition on the same information. This alignment makes the signed probability gap a token-level preference signal for sampled execution, with the report remaining available to guide later planning decisions.

\clearpage\scadreservepage
\subsection{Prefix Construction and Planning Credit}
\label{app:algorithm-tree}
Each input has its own tree, built after rollout collection from the ordered assignments and reports. Node identity includes the complete canonical parent prefix. Action values group the same subtask choice under a parent even when its reports lead to different children. Reports identify the resulting canonical histories, while terminal returns determine the relative planning value of each choice.

\begin{scadalgorithm}{Module 2: Prefix Construction and Planning Credit}{alg:planning-credit}
\begin{algorithmic}[1]
\Require Rollout paths $\{(d_{ik},r_{ik})_k\}_i$, terminal returns $\{R_i\}_i$, planning spans
\Ensure Planning-token tree credit $B_{it}$ with zero credit outside recorded spans
\State Deduplicate rollout identities; initialize one root and empty visit sets for input $x$
\State Normalize and sort unique assignments; greedily cluster at cosine threshold $0.83$
\State Canonicalize reports within each subtask cluster using typed matching below
\For{each distinct rollout $i$}
  \State $p\gets\operatorname{root}(x)$
  \For{each ordered assignment--report pair $(d_{ik},r_{ik})$}
    \State $z\gets\operatorname{cluster}(d_{ik})$; add $i$ to $\mathcal I(p,z)$ once
    \State $p\gets\operatorname{node}(p,z,\kappa_z(r_{ik}))$; record the parent and child identities
  \EndFor
  \State Add the terminal marker \texttt{STOP} under $p$ and record $i$ in $\mathcal I(p,\texttt{STOP})$
\EndFor
\State Compute input-level mean $\mu_x$ and sample standard deviation $\sigma_x$ of $R_i$
\For{each parent $p$ with observed choices $\mathcal A(p)$}
  \State $n_{pa}\gets|\mathcal I(p,a)|$ and $Q(p,a)\gets\operatorname{mean}_{i\in\mathcal I(p,a)}R_i$ for each $a$
  \State $N_p\gets\sum_a n_{pa}$; $\bar Q_p\gets\sum_a n_{pa}Q(p,a)/N_p$
  \State $V(p)\gets(N_p\bar Q_p+\nu\mu_x)/(N_p+\nu)$ with prior count $\nu=2$
  \State Compute $s_p$ from action values repeated by visit count, as in Eq.~\ref{eq:tree-scale}
  \State Set $\widetilde\sigma_p$ to the first value above $\epsilon$ in $(s_p,\sigma_x,1)$
  \State $A^{\rm tree}(p,a)\gets\operatorname{clip}((Q(p,a)-V(p))/\widetilde\sigma_p,-b,b)$ for each $a$
\EndFor
\State Initialize $B_{it}\gets0$; add each decision's tree credit on its recorded planning span
\State Apply the generated-token mask and \Return $B_{it}$ with the frozen tree metadata
\end{algorithmic}
\end{scadalgorithm}

\noindent\textbf{Typed report matching.} Reports first lose their return wrapper and undergo Unicode, case, punctuation, and whitespace normalization. Abstention cues in the first 200 characters take precedence. Otherwise, reports longer than six normalized words remain whole for entity matching; short reports are tested as booleans, dates, numbers, then entities. Types must agree. Boolean and numeric keys match exactly; a year-only date may match a date in that year. Entity matching uses normalized lexical rules, then embedding cosine similarity of at least $0.90$. Lexical rules include token-order invariance, supported name variants, and string similarity. When numeric parsing is ambiguous, entity matching compares the full normalized report with the same lexical and semantic rules.

\noindent\textbf{Determinism and stopping.} Unique texts and candidate keys are sorted before greedy matching, and the first compatible leader defines each cluster. Different parent prefixes remain distinct. Every stored path has a terminal marker, including interrupted paths; only a recorded generated \texttt{finish} span can receive its \texttt{STOP} credit. Duplicate sequence rows do not add return observations, and tree metadata is fixed before padding and batch reordering, then retained throughout policy optimization.

\noindent\textbf{Computational cost.} With $U$ unique assignments, $W$ unique report keys, and embedding dimension $d_e$, greedy semantic comparisons take at most $O((U^2+W^2)d_e)$ arithmetic operations, excluding text encoding and lexical matching. Return aggregation is linear in recorded path visits with hash-indexed sets; span projection touches the recorded decision intervals. No additional environment rollout or learned value model is required to estimate these planning advantages on the collected batch.

\noindent\textbf{Credit at sparse prefixes.} For a parent with one observed choice, prior shrinkage supplies a credit offset relative to the input-level mean. With multiple choices, the shared baseline also preserves their pre-clipping value differences, supporting comparisons among observed alternatives. The local scale uses sample standard deviation, with zero scale for a single visit. The input-level scale, then one, supplies the fallback for normalization across different visit counts and return distributions.

\clearpage\scadreservepage
\setlength{\parskip}{5pt}
\subsection{Signal Allocation and End-to-End Training}
\label{app:algorithm-training}
The training loop groups independent rollouts by task input, retains one terminal return per rollout, and allocates token advantages by planning or execution role. Planning receives signed outcome and tree credit. Execution receives the positive part of outcome credit and the signed teacher gap. The teacher, semantic encoder, stored contexts, and computed advantages remain fixed during each policy update, so gradients act only on the shared student policy parameters at the current training step.

\begin{scadalgorithm}{Module 3: Structured Signal Allocation and Policy Training}{alg:scad-training}
\begin{algorithmic}[1]
\Require Dataset $\mathcal D$, initial policy $\pi_\theta$, frozen teacher and encoder, $G$ rollouts
\Ensure Updated policy parameters and checkpoints for validation selection
\For{each training step}
  \State Freeze a rollout snapshot $\pi_{\rm old}\gets\pi_\theta$; sample a batch from $\mathcal D$
  \For{each task $(x,y^\star)$ in the batch}
    \State Collect $G$ independent rollouts and execution signals using Algorithm~\ref{alg:local-execution}
    \State Evaluate each submitted answer against $y^\star$ to obtain terminal return $R_i$
    \State Deduplicate rollout identities; compute $\mu_x$ and sample standard deviation $\sigma_x$
    \State $A_i^{\rm grp}\gets(R_i-\mu_x)/(\sigma_x+\epsilon)$ for each rollout $i$
    \State Compute planning-span credit $B_{it}$ with Algorithm~\ref{alg:planning-credit}
    \For{each generated token in the stored planning and execution sequences}
      \If{the token belongs to planning}
        \State $A_{it}\gets A_i^{\rm grp}+B_{it}$
      \Else
        \State $A_{it}\gets\max(A_i^{\rm grp},0)+\delta_{it}$
      \EndIf
    \EndFor
  \EndFor
  \State Form mask $m_{it}$ excluding prompts, observations, padding, and excluded sequences
  \State Optimize the token-averaged dual-clip objective in Eq.~\ref{eq:policy-update} with AdamW
  \State Evaluate and save checkpoints at the configured validation intervals
\EndFor
\State \Return the policy checkpoint selected using validation macro accuracy
\end{algorithmic}
\end{scadalgorithm}

\noindent\textbf{Optimization settings and masks.} The main runs use 60 updates, eight task inputs per batch, $G=8$, learning rate $10^{-6}$, and local growth budget $B=16{,}384$. Teacher gaps and tree advantages are clipped to $[-5,5]$, with unit signal weights. PPO uses ratio clip width $0.2$ and dual-clip factor $3$; the implementation also clamps the log probability ratio to $[-20,20]$ before exponentiation for numerical stability. The loss averages valid generated tokens, including generated reasoning and tool arguments. Prompt tokens and tool responses condition the policy but do not contribute direct loss terms.

\noindent\textbf{Text and multimodal scoring.} Text execution retains the question, parsed assignment, and local observations. Multimodal execution also retains the original pixels, expanded image-token identifiers, and matching visual positions for student and teacher scoring. The processor checks image-token counts before scoring. Image search supplies textual matches and source identifiers for subsequent page reading. Gemini-3.7-Flash compares the final answer with references, while distillation scores the sampled execution under the same task input and local evidence used by the student.

\noindent\textbf{Training cost and inference.} Collection costs depend on rollout decoding and tool access. The frozen teacher scores the collected batch, and execution-token scores supply the distillation signal; student optimization requires forward and backward passes. Local execution segments form the scoring inputs, whose lengths and count determine model computation. At inference, the shared policy uses greedy decoding with the original question and images, coordinating tool use through assignments and returned reports. Teacher scoring and tree-credit estimation operate during training.

\noindent\textbf{Group statistics and loss masks.} Each distinct rollout contributes one terminal return to the input-level statistics. Equal returns give zero outcome advantage and retain execution-side distillation. Generated-token masks select retained sequences, excluding padding and marked overlong or unfinished sequences. Shared rollout identities link planning and execution rows to the same outcome, assigning consistent task-level credit to the tokens used in each update of the shared policy.
\par\clearpage\endgroup

%% file: sections/benchmark_details.tex
\begingroup
\raggedbottom
\setlength{\parskip}{2.8pt}
\section{Benchmark Detail}
\label{app:benchmarks}
\label{app:resource-sources}
The 19 datasets contain 1,324 test questions: 814 text (525 ID, 289 OOD) and 510 multimodal (395 ID, 115 OOD). In Tables~\ref{tab:text-results}--\ref{tab:multimodal-results}, $\dagger$ marks seven OOD datasets absent from training. Bamboogle, PopQA, and FVQA contribute validation examples for checkpoint selection; GAIA, xbench, InfoSeek, and LiveVQA are used only for evaluation.

\subsection{Dataset Details}
\label{app:dataset-sources}
\noindent\textbf{Text ID (seven datasets).} BrowseComp-Plus \citep{chen2025browsecompplus} and WebWalkerQA \citep{wu-etal-2025-webwalker} contribute 120 and 80 questions on fact-finding and web traversal. HotpotQA (Hotpot) \citep{yang-etal-2018-hotpotqa}, 2WikiMultiHopQA (2Wiki) \citep{ho-etal-2020-constructing}, and MuSiQue (MuS) \citep{trivedi-etal-2022-musique} contribute 65 questions each on multi-step evidence composition. Natural Questions (NQ) \citep{kwiatkowski-etal-2019-natural} and TriviaQA (Trivia) \citep{joshi-etal-2017-triviaqa} each add 65 questions covering natural information needs and trivia.

\noindent\textbf{Text OOD (four datasets).} Bamboogle (Bamb.) \citep{press-etal-2023-measuring} contributes 65 multi-hop questions; PopQA \citep{mallen-etal-2023-trust} contributes 65 factual questions covering entities across popularity levels. GAIA \citep{mialon2024gaia} adds 59 selected text tasks combining practical reasoning with information access. The 100 xbench-DeepSearch (xbench) questions \citep{chen2025xbenchtrackingagentsproductivity} test extended research and evidence gathering.

\noindent\textbf{Multimodal ID (five datasets).} MMSearch-Plus \citep{tao2026mmsearchplus} contributes 105 questions linking fine-grained visual clues to facts through iterative search. Vision-DeepResearch Benchmark (VDR-Bench) \citep{zeng2026visiondeepresearchbenchmarkrethinkingvisual} contributes 135 questions on visual and textual fact-finding, and VisBrowse-Bench \citep{zhang2026visbrowsebenchbenchmarkingvisualnativesearch} contributes 65 on cross-modal verification. MMSearch \citep{jiang2024mmsearchbenchmarkingpotentiallarge} adds 55 multimodal search questions, and SimpleVQA \citep{Cheng_2025_ICCV} adds 35 questions connecting visual subjects to external knowledge, yielding 395 ID questions.

\noindent\textbf{Multimodal OOD (three datasets).} FactualVQA (FVQA) from MMSearch-R1 \citep{wu2025mmsearchr1incentivizinglmmssearch} contributes 35 questions linking images to factual knowledge. InfoSeek \citep{chen-etal-2023-pre-trained} contributes 35 questions about visually identified entities. LiveVQA \citep{NEURIPS2025_6c7c9811} adds 45 questions grounded in news imagery and time-sensitive knowledge. The resulting 115-question OOD pool covers factual, entity-centric, and news-based visual reasoning with original question--image inputs.

\noindent\textbf{Task records.} Each example retains its source dataset, instance identifier, question, and acceptable reference answers. Multimodal records also retain the original images, keeping the visual subject available during local execution. Reference answers are stored in evaluator metadata as lists that retain all supplied acceptable answer alternatives.

\subsection{Evaluation Details}
\label{app:benchmark-evaluation}
\noindent\textbf{Inference and grading.} The student uses greedy decoding independently of the distillation teacher, receiving the question and original images for multimodal tasks. Gemini-3.7-Flash scores final answers against the references. Text evaluation first accepts matches under the implemented answer-normalization rules, then invokes the judge for semantic equivalence. Multimodal evaluation supplies the answer alternatives directly to the same judge, accepting a match to any alternative. Equivalent wording and minor formatting differences are allowed; empty answers receive zero. With correctness $c_{dj}\in\{0,1\}$ for question $j$ in dataset $d$, final-answer accuracy (\%) and the macro average over $\mathcal S$ are
\begin{equation}
\operatorname{Acc}(d)=\frac{100}{n_d}\sum_{j=1}^{n_d}c_{dj},
\qquad
\operatorname{Avg}(\mathcal S)=\frac{1}{|\mathcal S|}\sum_{d\in\mathcal S}\operatorname{Acc}(d),
\label{eq:benchmark-accuracy}
\end{equation}
where $n_d$ is the number of evaluated questions in dataset $d$. The same binary answer outcome supplies the terminal training reward, linking task accuracy to the training supervision in Appendix~\ref{app:algorithms}.

\noindent\textbf{Aggregation and model selection.} Text Avg. averages all 11 datasets. Multimodal ID Avg. averages five ID datasets; All Avg. averages all eight. Figure accuracies use the same equal-dataset weighting. The ID/OOD macro averages combine as $(7A_{\rm ID}+4A_{\rm OOD})/11$ for text and $(5A_{\rm ID}+3A_{\rm OOD})/8$ for multimodal tasks. Training uses separate pools of 400 text and 347 multimodal examples. Checkpoints use validation macro accuracy, including the three validation-only OOD sources above. ID/OOD reflects training coverage independently of validation membership. RQ1--RQ2 cover both modalities; RQ3--RQ6 use text tasks (Section~\ref{sec:experimental-setup}).
\par\clearpage
\endgroup

%% file: sections/resource_sources.tex
\begingroup
\flushbottom
\setlength{\parskip}{2pt plus 3pt minus 0.25pt}
\makeatletter
\def\subsection{\@startsection{subsection}{2}{\z@}{-6pt}{3pt}{\normalsize\sc\raggedright}}
\makeatother
\interlinepenalty=10000
\predisplaypenalty=10000
\postdisplaypenalty=10000
\section{Baseline Detail}
\label{app:baselines}
\label{app:baseline-sources}
This appendix specifies the search implementations and hyperparameters of the baselines in Tables~\ref{tab:text-results}--\ref{tab:multimodal-results}. Methods share the models, training pools, tools, and evaluation protocol below while retaining their own context management and learning objectives.

\subsection{Shared Comparison Settings}
\label{app:baseline-protocol}
\noindent\textbf{Models and data.} Text methods use Qwen3-4B with frozen Qwen3-4B-Instruct-2507 as distillation teacher \citep{yang2025qwen3technicalreport}; Appendix~\ref{app:additional-results} substitutes a Qwen3-1.7B student. Multimodal methods use Qwen3-VL-2B and frozen Qwen3-VL-4B \citep{bai2025qwen3vltechnicalreport}. Training uses 400 text and 347 multimodal questions, including demonstrations, supervised initialization, and skill construction. SDAR instead obtains targets from a privileged-context branch of the student. Text training comprises 200 BrowseComp-Plus, 100 WebWalkerQA, and 20 each from NQ, TriviaQA, HotpotQA, 2WikiMultiHopQA, and MuSiQue; multimodal training comprises 102 MMSearch-Plus, 114 VDR-Bench, 45 VisBrowse-Bench, 52 MMSearch, and 34 SimpleVQA questions.

\noindent\textbf{Tools and masks.} Search agents share text search and page opening; multimodal agents also access image search and retain the original images. Methods preserve their flat, folded, hierarchical, or parallel interaction structures. Losses cover generated reasoning, tool arguments, reports, and answers. Prompts, retrieved observations, images, and padding provide context and are excluded from direct policy and distillation losses.

\noindent\textbf{Interaction budgets.} SCAD and search baselines share the same task budget. Worker and Planner budgets $B_W$ and $B_P$ measure transcript growth after initial messages. With at most $N_W$ Workers, the task budget is
\begin{equation}
B_{\rm total}=N_W B_W+B_P,\qquad B_W=16{,}384,\quad B_P=32{,}768.
\label{eq:baseline-budget}
\end{equation}
Training and validation use $N_W=6$, giving $131{,}072$ tokens. Text testing uses $N_W=10$ and $196{,}608$ tokens; multimodal testing retains $N_W=6$ and $131{,}072$ tokens. Flat agents use the same task budget across context continuations. Tree-GRPO reuses prefixes, charging new expansions against $G B_{\rm total}$ per question for up to $G=8$ terminal paths.

\noindent\textbf{Context continuation.} Flat OPD-family agents summarize at 80\% of a 32,768-token window. Summaries retain findings, source identifiers, unresolved checks, and next actions within 1,024 tokens. The question stays fixed; the latest interaction is retained when space permits. Continuation consumes the remaining rollout allowance. Folded methods return subtask reports; MemOPD updates compact memory between calls. Training scores actions under their original sampling contexts.

\noindent\textbf{Optimization and evaluation.} Unless specified otherwise, online training uses 60 updates, eight prompts with eight rollouts each, and one optimization pass per batch. AdamW uses learning rate $10^{-6}$, weight decay $0.01$, and gradient clipping $1.0$. Training temperature/top-$p$ are $1.0/1.0$; evaluation is greedy. GRPO-style ratio clipping is $0.2$, with dual-clip factor $3$. Checkpoints use validation macro accuracy; three seeds share splits, with mean and sample standard deviation reported. Gemini-3.7-Flash grades answers. Training may include the reward shaping below; evaluation uses binary correctness and dataset macro averages (Appendix~\ref{app:benchmark-evaluation}), with OOD defined by absence from training; Bamboogle, PopQA, and FVQA also contribute to validation.

\subsection{Reference Models and Supervised Fine-Tuning}
\noindent\textbf{Vanilla.} The pretrained student directly answers the question and original images, when present, without planning or retrieval. Greedy decoding allows 4,096 generated tokens. No task-specific updates are applied; repeated evaluations retain the same checkpoint and grading protocol. This direct-generation reference differs from the scaffolded, untrained student at the start of RQ1.

\noindent\textbf{Teacher.} The frozen teacher uses SCAD's Planner--Worker interface, tools, and modality-specific test allowance with greedy decoding. Its table row measures generated final-answer accuracy under the shared judge and aggregation. During distillation, the same model instead scores token probabilities on student-generated responses.

\noindent\textbf{SFT.} Fixed teacher trajectories from the shared training pools include intermediate responses, tool calls, and answers; only response tokens receive cross-entropy loss. The student retains the learned tool-use protocol at evaluation. \emph{Settings.} AdamW uses $5\times10^{-6}$, batch size eight, two epochs, and 10\% warmup followed by linear decay. Long demonstrations use causally valid invocation segments of at most 32,768 tokens.

\subsection{Reinforcement-Learning Baselines}
\noindent\textbf{GRPO} \citep{shao2024deepseekmathpushinglimitsmathematical}. A flat search agent samples complete interactions and assigns each generated token its trajectory's normalized terminal reward, retaining signed advantages. \emph{Settings.} The shared $8\times8$, 60-update recipe uses reference-KL coefficient $10^{-3}$ toward the frozen initial student and zero entropy bonus, without external distillation or a learned critic.

\noindent\textbf{FoldGRPO} \citep{sun2026scaling}. Branch-and-return operations execute subtasks in separate contexts and fold reports back into the parent; nested branching is disabled. Parent and branch responses are jointly optimized with signed terminal credit. \emph{Settings.} Worker/Planner allowances follow Eq.~\ref{eq:baseline-budget}; optimization follows GRPO. Our outcome-only configuration disables the original scope/process-reward terms while retaining learned context folding.

\noindent\textbf{Tree-GRPO} \citep{ji2026tree}. The sampler branches at ReAct reasoning/tool-use steps, reuses prefixes, and propagates terminal returns through the resulting tree. \emph{Settings.} Binary expansion over three branching levels yields up to eight terminal paths, continued to an answer within the shared budget. Learning rate, clipping, and reference-KL weight follow GRPO; each sampled tree edge contributes once to the loss.

\noindent\textbf{HiPER} \citep{peng2026hiper}. A shared policy proposes subgoals and executes search actions. Subgoal segments define hierarchical advantages; a value head estimates planning and execution baselines from terminal correctness. Planning credit propagates between subgoals, while execution credit propagates within each segment, aligning policy updates with the observed hierarchy. \emph{Settings.} Discount $\gamma=1$, GAE $\lambda=0.95$, actor/critic learning rates $10^{-6}$ and $5\times10^{-6}$, and value-loss weight $0.5$. Both levels retain signed advantages and the shared sampling budget.

\subsection{On-Policy Distillation and Hybrid Methods}
\noindent\textbf{Teacher inputs.} Teachers score student responses under recorded prefixes containing the question, earlier outputs, and observations, or reconstructed memory for MemOPD. Distillation averages valid response positions. The initial-policy reference used for RL regularization controls drift from initialization; the distillation teacher supplies the transfer signal. Their KL coefficients therefore govern separate objectives.

\noindent\textbf{OPD} \citep{agarwal2024onpolicy,gu2024minillm}. The student generates search trajectories, and the teacher scores sampled response tokens under matching prefixes. The log-probability gap supplies dense guidance, with summary-enabled continuation. \emph{Settings.} Distillation weight $1.0$, gap clipping $[-5,5]$, terminal-reward weight zero, and 60 updates without an additional reference-KL penalty.

\noindent\textbf{SDAR} \citep{lu2026selfdistilledagenticreinforcementlearning}. A detached student branch receives a training-only skill summary; the acting student sees ordinary search context. Semantic retrieval selects one summary of at most 512 tokens from training demonstrations. \emph{Settings.} Self-distillation weight $0.1$ and sigmoid sharpness $1.0$ give $g_t=\operatorname{sigmoid}(\operatorname{sg}[\Delta_t])$, emphasizing positive teacher--student log-probability gaps. The gate preserves signed GRPO credit; evaluation removes privileged skill context.

\noindent\textbf{ATOD} \citep{tan2026atodannealedturnawareonpolicy}. Signed GRPO credit is combined with teacher guidance weighted by turn disagreement and uncertainty, normalized within each trajectory and combined by soft-OR. Reweighting affects only distillation. \emph{Settings.} With $u(s)=\min(s/32,1)$, OPD/RL weights are $1-0.9u(s)$ and $1+u(s)$, moving from $1.0/1.0$ to $0.1/2.0$ in 32 updates. Continuous turn weights activate for at least three turns and maximum disagreement $\geq0.01$. Actor learning rate is $10^{-6}$ with six warmup updates. A teacher anchor remains after annealing, whereas sequential OPD--RL schedules \citep{li2026sequentialbeatsjointinterplay} separate imitation and reward optimization into successive phases.

\noindent\textbf{TCOD} \citep{wang2026exploring}. The forward-to-backward curriculum expands the student-controlled prefix, with teacher scoring within the active horizon. \emph{Settings.} $k(s)=\min(32,2+\lfloor30s/48\rfloor)$ grows from two to 32 turns over 48 updates, followed by 12 updates at full training depth. Distillation weight is one and terminal-reward weight is zero. Evaluation removes the curriculum and uses the complete interaction allowance with summary support.

\noindent\textbf{MemOPD} \citep{liu2026memopdonpolicydistillationmemory}. Training reconstructs each invocation's input, response, positions, and causal visibility after memory rewriting. Retained-text copies provide context; the original sampled actions receive teacher guidance and PPO credit. \emph{Settings.} Memory cap 1,024 tokens, full-vocabulary reverse-KL weight $0.1$, actor/critic learning rates $10^{-6}$ and $5\times10^{-6}$, $\gamma=1$, GAE $\lambda=0.95$, and value-loss weight $0.5$. Initialization uses one SFT epoch on teacher trajectories at $5\times10^{-6}$ with batch size eight.

\subsection{Multimodal Search and Distillation}
\noindent\textbf{MMSearch-R1} \citep{wu2025mmsearchr1incentivizinglmmssearch}. Multi-turn GRPO learns when to answer directly and when to retrieve visual or textual evidence. The policy chooses one action per round and integrates returned evidence into subsequent reasoning. We retain its accuracy, search-cost, and formatting reward components, using the shared correctness judge. \emph{Settings.} With correctness $c$, format indicator $f$, and $\rho=0.9$ if search is used ($1$ otherwise), the reward is $0.9c\rho+0.1f$. The optimizer follows GRPO; retrieved content is masked, and image search uses the common service.

\noindent\textbf{VSearcher} \citep{zhang2026vsearcherlonghorizonmultimodalsearch}. We adapt its reinforcement-learning stage to long-horizon multimodal search using the shared Qwen3-VL-2B student and multimodal training pool. The policy retains original image inputs and gathers visual and textual evidence through multi-round tool use. The student samples its own trajectories for terminal-reward GRPO, with signed advantages applied to generated responses. \emph{Settings.} Online training uses 60 updates at $10^{-6}$, with eight prompts and eight rollouts per prompt. RL uses a $10^{-3}$ reference-KL coefficient and no external teacher-scoring loss, retaining the common inference and evidence budgets.

\noindent\textbf{VOLD} \citep{Bousselham_2026_CVPR}. Joint GRPO and OPD combine outcome rewards with teacher guidance for image-grounded search. We use the shared Qwen3-VL teacher--student pair and multimodal training pool. Both models condition on the original images and student search history. Reward-guided masking applies distillation to unsuccessful trajectories, while GRPO trains on all trajectories. \emph{Settings.} Online training uses learning rate $10^{-6}$, eight prompts with eight rollouts each, and sampled-token $k_2$ distillation with coefficient $0.1$. The vision tower remains frozen, and language-policy parameters are optimized over the shared 60-update budget.

\noindent\textbf{HyperEyes} \citep{li2026hypereyesdualgrainedefficiencyawarereinforcement}. Parallel grounded search combines trajectory-level efficiency rewards with token-level correction. Multiple queries may share a round; both rounds and individual calls enter the efficiency reward. A frozen Qwen3-VL-4B teacher supplies OPD only on failed rollouts. \emph{Settings.} Online training uses 60 updates with the shared optimizer and an OPD coefficient of $0.05$. TRACE uses redundancy tolerance $1.5$, redundancy penalty $-0.1$, format penalty $-0.5$, and reward intervals $[0.05,0.20]$ for efficient and $[-0.10,-0.02]$ for redundant successful searches. Up to four queries execute concurrently; all queries count against the shared budget.

\noindent\textbf{Online comparison protocol.} VSearcher, VOLD, and HyperEyes start from the shared Qwen3-VL-2B checkpoint and use 60 online updates on the same multimodal training pool. Their search structures and learning objectives retain method-specific behavior under shared rollout budgets. Checkpoint selection and evaluation follow the common validation and test protocol.

\subsection{SCAD and Controlled Comparisons}
\label{app:implementation-sources}
\noindent\textbf{SCAD.} Local Workers preserve the question, assignment, and subtask observations for teacher scoring. The Planner receives signed terminal and prefix-tree credit; Workers receive positive terminal credit and signed local OPD. Training uses 60 updates, $8\times8$ rollouts, and AdamW at $10^{-6}$; all three signal weights are one, with teacher/tree clipping at five. Semantic matching uses Qwen3-Embedding-4B \citep{zhang2025qwen3embeddingadvancingtext}, subtask/report thresholds $0.83/0.90$, and prior count $\nu=2$. The verl/HybridFlow implementation \citep{10.1145/3689031.3696075} and full update procedure are described in Appendix~\ref{app:algorithms}.

\noindent\textbf{Component and context controls.} RQ2 holds the Planner--Worker architecture fixed and adds local OPD, tree credit, and structured allocation to local GRPO in that order. RQ3 uses RQ2's second stage before tree credit and positive-only allocation. Its adapted ATOD keeps full-history teacher supervision and its own GRPO signals; the main table uses standalone ATOD. RQ4 changes only planning credit, including a control multiplying normalized no-tree Planner advantages by $1.6$. RQ5 changes GRPO/OPD assignment within the shared framework; Full-E keeps SCAD planning and uses signed GRPO plus OPD for execution.

\noindent\textbf{Efficiency controls.} RQ6(a) holds the Planner--Worker framework fixed across all three methods and counts only teacher-supervised response tokens. In RQ6(b), GRPO + OPD uses the flat search interface with unit weights for signed terminal credit and teacher guidance, sharing the online training recipe and frozen teacher with ATOD and SCAD. RQ5's Full GRPO + OPD instead uses the Planner--Worker framework with planning tree credit. RQ6 averages three runs; panel (b) excludes Gemini-3.7-Flash judging from 60-step timing.
\par\pagebreak
\endgroup

%% file: sections/case_study.tex
\begingroup
\raggedbottom
\setlength{\parskip}{3pt}
\setlength{\emergencystretch}{1em}
\definecolor{caseBlue}{RGB}{43,89,164}
\definecolor{caseTeal}{RGB}{0,112,110}
\definecolor{caseRed}{RGB}{170,49,55}
\definecolor{casePurple}{RGB}{92,66,137}
\definecolor{caseShade}{RGB}{244,240,250}
\definecolor{caseMuted}{RGB}{91,96,104}
\newcommand{\caseaction}[1]{\textcolor{caseBlue}{\texttt{#1}}}
\newcommand{\casereport}{\textcolor{caseTeal}{\texttt{report}}}
\newcommand{\casefinish}{\textcolor{caseBlue}{\texttt{finish}}}
\newcommand{\caseblockgap}{\par\vspace{3pt}}

\section{Case Study}
\label{app:case-study}
\begin{minipage}{\textwidth}
\makeatletter\def\@captype{table}\makeatother
\setlength{\abovecaptionskip}{3pt}
\setlength{\belowcaptionskip}{5pt}
\caption{Three dependent subtasks in a VDR-Bench training rollout of SCAD (Qwen3-VL-2B). Completed branches and reports are condensed; the terminal reward is $R=1$.}
\label{tab:case-study}

\begin{tcolorbox}[colback=black!2,colframe=black!25,boxrule=0.4pt,
  arc=1pt,left=6pt,right=6pt,top=5pt,bottom=5pt,
  before skip=0pt,after skip=5pt]
\begin{minipage}[c]{0.35\linewidth}
\includegraphics[width=\linewidth]{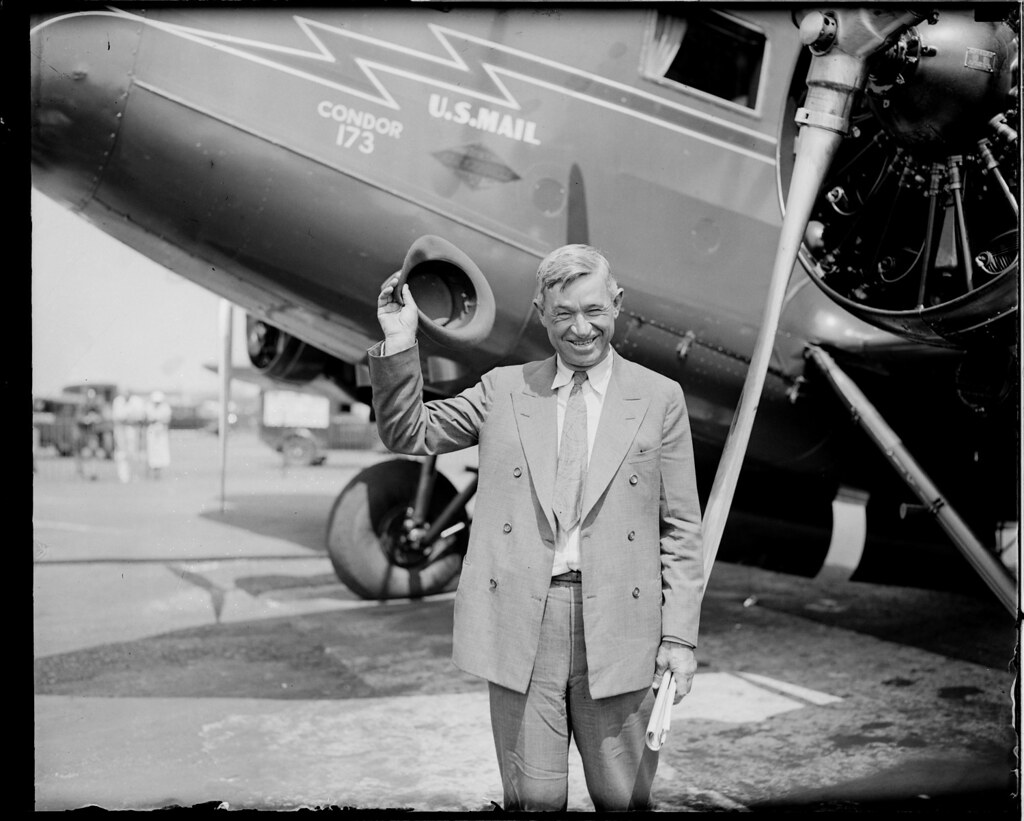}
\end{minipage}\hfill
\begin{minipage}[c]{0.62\linewidth}
\fontsize{10}{11.5}\selectfont
\textbf{Question.} The man in the foreground had a son who starred in a radio series that aired from 1953 to 1954. What was the genre of this series, and what was the name of the 1952 film that preceded it?
\par\vspace{4pt}
\textbf{Ground truth:} Light drama;\par
\emph{The Story of Will Rogers}.
\end{minipage}
\par\vspace{3pt}
{\fontsize{7.5}{9}\selectfont\color{caseMuted}Boston Public Library, Leslie Jones Collection.}
\end{tcolorbox}

\setlength{\tabcolsep}{6pt}
\renewcommand{\arraystretch}{1}
\fontsize{10}{11.5}\selectfont
\begin{tabularx}{\linewidth}{>{\raggedright\arraybackslash}p{0.18\linewidth}X}
\toprule
\textbf{Stage} & \textbf{Subtask and returned evidence} \\
\midrule
\textcolor{casePurple}{\textbf{Branch 1}}\par
Visual identity
&
\caseaction{identify person}\; Identify the foreground man and his associations with radio or film.
\caseblockgap
\casereport\; Identifies \textbf{Will Rogers}, but leaves his son's 1953--1954 series unresolved and includes an unverified 1952 title.
\\[5pt]
\midrule
\rowcolor{caseShade}
\textcolor{casePurple}{\textbf{Branch 2}}\par
Family link\par
and radio series
&
\caseaction{trace son's series}\; Find \textbf{Will Rogers' son's} radio series from 1953--1954 and confirm its genre.
\caseblockgap
\casereport\; \textbf{Will Rogers Jr.} starred in \textbf{Rogers of the Gazette}, a \textbf{drama}; the report also links it to \emph{The Story of Will Rogers} (1952).
\\[5pt]
\midrule
\textcolor{casePurple}{\textbf{Branch 3}}\par
Film verification
&
\caseaction{verify precursor}\; Confirm the exact title of the 1952 film preceding \textbf{Rogers of the Gazette}.
\caseblockgap
\casereport\; Confirms \textbf{The Story of Will Rogers} as the film that inspired the radio series, with Will Rogers Jr. carrying his father's persona into the radio role.
\\[5pt]
\midrule
\rowcolor{caseShade}
\textcolor{casePurple}{\textbf{SCAD answer}}\par
\textcolor{caseTeal}{\bfseries\mbox{$R=1$\enspace Correct}}
&
\casefinish\; \textcolor{caseTeal}{\textbf{Drama; The Story of Will Rogers (1952).}}\par
Three returned reports connect the photograph to both requested facts.
\\[5pt]
\bottomrule
\end{tabularx}
\end{minipage}

\noindent\textbf{Three distinct intermediate results.} SCAD identifies the person, traces his son's radio series and its genre, and verifies the preceding film. The Planner reads each report and writes the relevant names into subsequent subtask instructions. The final branch verifies the film connection before the Planner submits the complete answer. The three branches thus link visual identification, relational retrieval, and film verification through separate local executions.
\par
\endgroup

%% file: sections/limitations_future_work.tex
\begingroup
\raggedbottom
\setlength{\parskip}{4pt}
\section{Limitations and Future Work}
\label{app:limitations}

\noindent\textbf{Limitations.} SCAD is evaluated with compact Qwen students on text and image-based search tasks. Local distillation depends on the selected teacher's competence on sampled subtasks. Prefix-tree credit draws on semantically matched subtask--report histories, so its comparison quality depends on matching accuracy and the number of observed alternatives. Diverse prefixes may yield sparse sibling comparisons, for which the current method uses prior shrinkage. Worker budgets and signal weights remain fixed across tasks, leaving room to adapt to subtask difficulty and feedback quality.

\noindent\textbf{Future Work.} We aim to extend SCAD to larger backbones and a broader range of interactive tasks, while studying teacher selection by local competence. Adaptive rollout sampling and learned history matching could increase useful comparisons within the prefix tree. A complementary direction is to allocate Worker budgets and teacher supervision according to subtask difficulty and expected learning benefit. Adapting these allocations together with training signal weights could improve the balance between task accuracy and computation across tasks with different evidence requirements.
\par
\endgroup